%% file: main.tex
\documentclass[runningheads]{llncs}

\usepackage{eccv}
\usepackage[toc,page]{appendix}

\usepackage{eccvabbrv}

\usepackage{graphicx}
\usepackage{booktabs}

\usepackage[accsupp]{axessibility}  
\usepackage{multirow}
\usepackage{color}

\definecolor{c1}{HTML}{ffb61e}
\usepackage[ruled]{algorithm2e}

\usepackage{hyperref}

\usepackage{orcidlink}

\usepackage{eccvabbrv}

\usepackage{graphicx}
\usepackage{booktabs}
\usepackage{footnote}
\usepackage{lipsum}

\newcommand\blfootnote[1]{%
  \begingroup
  \renewcommand\thefootnote{}\footnote{#1}%
  \addtocounter{footnote}{-1}%
  \endgroup
}

\usepackage[accsupp]{axessibility}  
\usepackage[ruled]{algorithm2e}
\usepackage{orcidlink}
\definecolor{c1}{HTML}{ffb61e}
\definecolor{c2}{HTML}{057748}
\begin{document}

\title{GaussianFlow: Splatting Gaussian Dynamics for 4D Content Creation} 

\titlerunning{GaussianFlow}

\author{Quankai Gao\inst{1,2} \and Qiangeng Xu\inst{2} \and
Zhe Cao\inst{2} \and Ben Mildenhall\inst{2} \and Wenchao Ma\inst{3} \and Le Chen\inst{4}\and Danhang Tang\inst{2} \and Ulrich Neumann\inst{1}}

\authorrunning{Quankai Gao et al.}

\institute{University of Southern California  \\
\and
Google\\
\and
Pennsylvania State University\\
\and
Max Planck Institute for Intelligent Systems\\}


\maketitle

\blfootnote{\leftline{$^\dagger$ Contact for paper details: quankaig@usc.edu, qiangenx@google.com.}}

\input{teaser}

\begin{abstract}
    Creating 4D fields of Gaussian Splatting from images or videos is a challenging task due to its under-constrained nature. While the optimization can draw photometric reference from the input videos or be regulated by generative models, directly supervising Gaussian motions remains underexplored. In this paper, we introduce a novel concept, Gaussian flow, which connects the dynamics of 3D Gaussians and pixel velocities between consecutive frames. The Gaussian flow can be efficiently obtained by splatting Gaussian dynamics into the image space. This differentiable process enables direct dynamic supervision from optical flow. Our method significantly benefits 4D dynamic content generation and 4D novel view synthesis with Gaussian Splatting, especially for contents with rich motions that are hard to be handled by existing methods. The common color drifting issue that happens in 4D generation is also resolved with improved Guassian dynamics. Superior visual quality on extensive experiments demonstrates our method's effectiveness. Quantitative and qualitative evaluations show that our method achieves state-of-the-art results on both tasks of 4D generation and 4D novel view synthesis. \\
    Project page: \href{https://zerg-overmind.github.io/GaussianFlow.github.io/}{https://zerg-overmind.github.io/GaussianFlow.github.io/}  
    
  
  
  
\end{abstract}


\section{Introduction}
\label{sec:intro}
4D dynamic content creation from monocular or multi-view videos has garnered significant attention from academia and industry due to its wide applicability in virtual reality/augmented reality, digital games, and movie industry. Studies  ~\cite{li2022neural,pumarola2021d,park2021nerfies,park2021hypernerf} model 4D scenes by 4D dynamic Neural Radiance Fields (NeRFs) and optimize them based on input multi-view or monocular videos. Once optimized, the 4D field can be viewed from novel camera poses at preferred time steps through volumetric rendering. A more challenging task is generating 360 degree 4D content based on uncalibrated monocular videos or synthetic videos generated by text-to-video or image-to-video models. Since the monocular input cannot provide enough multi-view cues and unobserved regions are not supervised due to occlusions, studies \cite{singer2023text,jiang2023consistent4d,zhao2023animate124} optimizes 4D dynamic NeRFs by leveraging generative models to create plausible and temporally consistent 3D structures and appearance. The optimization of 4D NeRFs requires volumetric rendering which makes the process time-consuming. And real-time rendering of optimized 4D NeRFs is also hardly achieved without special designs. A more efficient alternative is to model 4D Radiance Fields by 4D Gaussian Splatting (GS) \cite{wu20234d,luiten2023dynamic}, which extends 3D Gaussian Splatting \cite{kerbl20233d} with a temporal dimension. Leveraging the efficient rendering of 3D GS, the lengthy training time of a 4D Radiance Field can be drastically reduced \cite{yang2023real,ren2023dreamgaussian4d} and rendering can achieve real-time speed during inference.

The optimization of 4D Gaussian  fields takes photometric loss as major supervision. As a result, the scene dynamics are usually under-constraint. Similarly to 4D NeRFs \cite{li2023dynibar,park2021nerfies,pumarola2021d}, the radiance properties and the time-varying spatial properties (location, scales, and orientations) of Gaussians are both optimized to reduce the photometric Mean Squared Error (MSE) between the rendered frames and the input video frames. The ambiguities of appearance, geometry, and dynamics have been introduced in the process and become prominent with sparse-view or monocular video input. Per-frame Score Distillation Sampling (SDS) \cite{tang2023dreamgaussian} reduces the appearance-geometry ambiguity to some extent by involving multi-view supervision in latent domain. However, both monocular photometric supervision and SDS supervision do not directly supervise scene dynamics. 

To avoid temporal inconsistency brought by fast motions, Consistent4D~\cite{jiang2023consistent4d} leverages a video interpolation block, which imposes a photometric consistency between the interpolated frame and generated frame, at a cost of involving more frames as pseudo ground truth for fitting. Similarly, AYG~\cite{ling2023align} uses text-to-video diffusion model to balance motion magnitude and temporal consistency with a pre-set frame rate.
4D NeRF model \cite{li2023dynibar} has proven that optical flows on reference videos are strong motion cues and can significantly benefit scene dynamics. However, for 4D GS, connecting 4D Gaussian motions with optical flows has following two challenges. First, a Gaussian's motion is in 3D space, but it is its 2D splat that contributes to rendered pixels. Second, multiple 3D Gaussians might contribute to the same pixel in rendering, and each pixel's flow does not equal to any one Gaussian's motion.


To deal with these challenges, we introduce a novel concept, Gaussian flow, bridging the dynamics of 3D Gaussians and pixel velocities between consecutive frames. Specifically, we assume the optical flow of each pixel in image space is influenced by the Gaussians that cover it. The Gaussian flow of each pixel is considered to be the weighted sum of these Gaussian motions in 2D. To obtain the Gaussian flow value on each pixel without losing the speed advantage of Gaussian Splatting, we splat 3D Gaussian dynamics, including scaling, rotation, and translation in 3D space, onto the image plane along with its radiance properties. As the whole process is end-to-end differentiable, the 3D Gaussian dynamics can be directly supervised by matching Gaussian flow with optical flow on input video frames. We apply such flow supervision to both 4D content generation and 4D novel view synthesis to showcase the benefit of our proposed method, especially for contents with rich motions that are hard to be handled by existing methods. The flow-guided Guassian dynamics also resolve the color drifting artifacts that are commonly observed in 4D Generation. We summarize our contributions as follows:

\begin{itemize}
    
    \item {We introduce a novel concept, Gaussian flow, that first time bridges the 3D Gaussian dynamics to resulting pixel velocities. Matching Gaussian flows with optical flows, 3D Gaussian dynamics can be directly supervised.}

    \item {The Gaussian flow can be obtained by splatting Gaussian dynamics into the image space. Following the tile-based design by original 3D Gaussian Splatting, we implement the dynamics splatting in CUDA with minimal overhead. The operation to generate dense Gaussian flow from 3D Gaussian dynamics is highly efficient and end-to-end differentiable.}
        
    \item {With Gaussian flow to optical flow matching, our model drastically improves over existing methods, especially on scene sequences of fast motions. Color drifting is also resolved with our improved Gaussian dynamics.}
\end{itemize}





\section{Related Works}
\subsubsection{3D Generation.} 3D generation has drawn tremendous attention with the progress of various 2D or 3D-aware diffusion models~\cite{liu2023zero,rombach2022high,shi2023mvdream,liu2023syncdreamer} and large vision models~\cite{radford2021learning,jun2023shap,nichol2022point}. Thanks to the availability of large-scale multi-view image datasets~\cite{deitke2023objaverse,yu2023mvimgnet,downs2022google}, object-level multi-view cues can be encoded in generative models and are used for generation purpose. Pioneered by DreamFusion~\cite{poole2022dreamfusion} that firstly proposes Score Distillation Sampling (SDS) loss to lift realistic contents from 2D to 3D via NeRFs, 3D content creation from text or image input has flourished. This progress includes approaches based on online optimization~\cite{tang2023dreamgaussian,lin2023magic3d,wang2024prolificdreamer,raj2023dreambooth3d} and feedforward methods~\cite{hong2023lrm,liu2023one,liu2024one,xu2023dmv3d,wang2023pf} with different representations such as NeRFs~\cite{mildenhall2021nerf}, triplane~\cite{chan2022efficient,chen2022tensorf,gao2023strivec} and 3D Gaussian Splatting~\cite{kerbl20233d}. 3D generation becomes more multi-view consistent by involving multi-view constraints~\cite{shi2023mvdream} and 3D-aware diffusion models~\cite{liu2023zero} as SDS supervision. Not limited to high quality rendering, some works~\cite{sun2023dreamcraft3d,long2023wonder3d} also explore enhancing the quality of generated 3D geometry by incorporating normal cues.

\subsubsection{4D Novel View Synthesis and Reconstruction.} By adding timestamp as an additional variable, recent 4D methods with different dynamic representations such as dynamic NeRF~\cite{park2021nerfies,park2021hypernerf,li2021neural,wang2023flow,li2022neural,tretschk2021non,gao2021dynamic}, dynamic triplane~\cite{fridovich2023k,cao2023hexplane,shao2023tensor4d} and 4D Gaussian Splatting~\cite{wu20234d,yang2023real} are proposed to achieve high quality 4D motions and scene contents reconstruction from either calibrated multi-view or uncalibrated RGB monocular video inputs. There are also some works~\cite{newcombe2011kinectfusion,newcombe2015dynamicfusion,zollhofer2014real} reconstruct rigid and non-rigid scene contents with RGB-D sensors, which help to resolve 3D ambiguities by involving depth cues.
Different from static 3D reconstruction and novel view synthesis, 4D novel view synthesis consisting of both rigid and non-rigid deformations is notoriously challenging and ill-posed with only RGB monocular inputs. Some progress~\cite{li2021neural,gao2021dynamic,tretschk2021non,wang2021neural} involve temporal priors and motion cues (e.g. optical flow) to better regularize temporal photometric consistency and 4D motions. One of recent works~\cite{wang2023flow} provides an analytical solution for flow supervision on deformable NeRF without inverting the backward deformation function from world coordinate to canonical coordinate. Several works~\cite{yang2021lasr,yang2021viser,yang2023reconstructing,yang2023ppr} explore object-level mesh recovery from monocular videos with optical flow. 
\subsubsection{4D Generation.} Similar to 3D generation from text prompts or single images, 4D generation from text prompts or monocular videos also relies on frame-by-frame multi-view cues from pre-trained diffusion models. Besides, 4D generation methods yet always rely on 
either video diffusion models or video interpolation block to ensure the temporal consistency. Animate124~\cite{zhao2023animate124}, 4D-fy~\cite{bahmani20234d} and one of the earliest works~\cite{singer2023text} use dynamic NeRFs as 4D representations and achieve temporal consistency with text-to-video diffusion models, which can generate videos with controlled frame rates. Instead of using dynamic NeRF, Align Your Gaussians~\cite{ling2023align} and DreamGaussian4D~\cite{ren2023dreamgaussian4d} generate vivid 4D contents with 3D Gaussian Splatting, but again, relying on text-to-video diffusion model for free frame rate control. Without the use of text-to-video diffusion models, Consistent4D~\cite{jiang2023consistent4d} achieves coherent 4D generation with an off-the-shelf video interpolation model~\cite{huang2022real}. Our method benefits 4D Gaussian representations by involving flow supervision and without the need of specialized temporal consistency networks.    

\vspace{-2mm}
\section{Methodology}
\begin{figure}[tb]
  \centering
  \includegraphics[height=5.9cm]{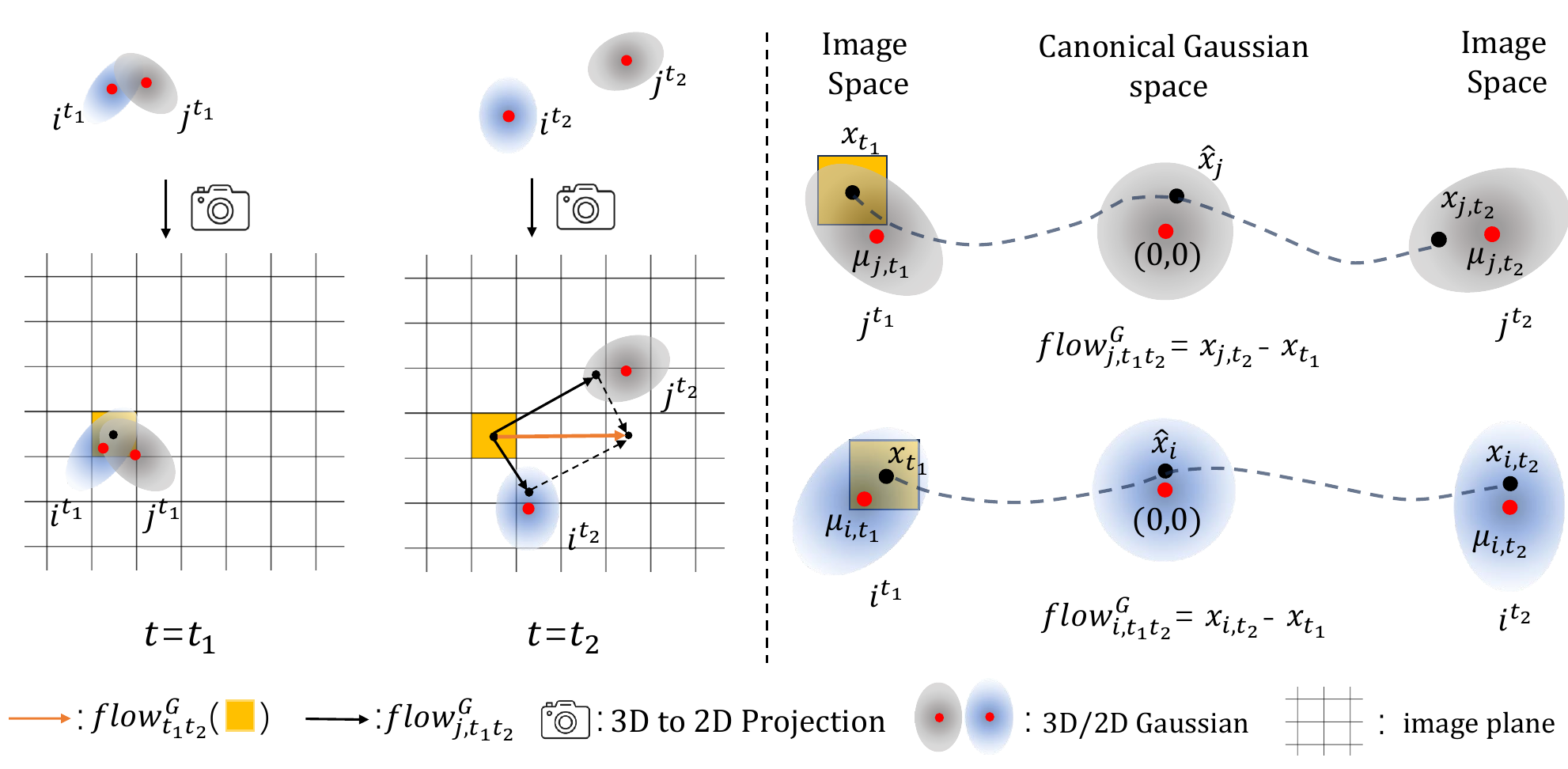}
  \vspace{-3mm}
  \caption{Between two consecutive frames, a pixel $x_{t_1}$ will be pushed towards $x_{t_1} \rightarrow x_{i,t_2}$ by the 2D Gaussian $i$'s motion $i^{t_1} \rightarrow i^{t_2}$. We can track $x_{t_1}$ in Gaussian $i$ by normalizing it to canonical Gaussian space as $\hat{x}_i$ and unnormalize it to image space to obtain $x_{i,t_2}$. Here, we denote this shift contribution from Gaussian $i$ as $flow^G_{i,t_1,t_2}$. The Gaussian flow $flow^G_{t_1,t_2}(x_{t_1})$ on pixel $x_{t_1}$ is defined as the weighted sum of the shift contributions from all Gaussians covering the pixel ($i$ and $j$ in our example). The weighting factor utilizes alpha composition weights. The Gaussian flow of the entire image can be obtained efficiently by splatting 3D Gaussian dynamics and rendering with alpha composition, which can be implemented similarly to the pipeline of the original 3D Gaussian Splatting \cite{kerbl20233d}.
  }
  \label{fig:example}
\end{figure}
To better illustrate the relationship between Gaussian motions and corresponding pixel flow in 2D images, we first recap the rendering process of 3D Gaussian Splatting and then investigate its 4D case.
\vspace{-2mm}
\subsection{Preliminary}

\subsubsection{3D Gaussian Splatting.} 
From a set of initialized 3D Gaussian primitives, 3D Gaussian Splatting aims to recover the 3D scene by minimizing photometric loss between input $m$ images $\{I\}_m$ and rendered images $\{I_r\}_m$. For each pixel, its rendered color $C$ is the weighted sum of multiple Gaussians' colors $c_i$ in depth order along the ray by point-based $\alpha$-blending as in Eq.~\ref{eq:rendering_eq},
\begin{equation}
 C = \sum^N_{i=1}T_i\alpha_ic_i,
  \label{eq:rendering_eq}
\end{equation}
with weights specifying as 
\begin{equation}
 \alpha_i = o_ie^{-\frac{1}{2}(\mathbf{x}-\boldsymbol{\mu}_i)^T\mathbf{\Sigma}_i^{-1}(\mathbf{x}-\boldsymbol{\mu}_i)} \quad  \text{and}  \quad  T_i=\sum^{i-1}_{j=1}(1-\alpha_i).
\label{eq:weights}
\end{equation}
where $o_i\in[0,1]$, $\boldsymbol{\mu}_i\in\mathbb{R}^{2\times1}$, and $\mathbf{\Sigma}_i\in\mathbb{R}^{2\times2}$ are the opacity, 2D mean, and 2D covariance matrix of $i$-th Gaussian, respectively. And $\mathbf{x}$ is the intersection between a pixel ray and $i$-th Gaussian. As shown in Eq.~\ref{eq:rendering_eq}, the relationship between a rendered pixel and 3D Gaussians is not bijective.

\subsubsection{3D Gaussian Splatting in 4D.} Modeling 4D motions with 3D Gaussian Splatting can be done frame-by-frame via either directly multi-view fitting~\cite{luiten2023dynamic} or moving 3D Gaussians with a time-variant deformation field~\cite{ling2023align,ren2023dreamgaussian4d} or parameterize 3D Gaussians with time~\cite{yang2023real}. While with monocular inputs, Gaussian motions are under-constrained because different Gaussian motions can lead to the same rendered color, and thus long-term persistent tracks are lost~\cite{luiten2023dynamic}. Though Local Rigidity Loss~\cite{luiten2023dynamic,ling2023align} is proposed to reduce global freedom of Gaussian motions, it sometimes brings severe problems due to poor or challenging initialization and lack of multi-view supervision. As shown in Fig.~\ref{fig:failure}, 3D Gaussians initialized with the skull mouth closed are hard to be split when the mouth open with Local Rigidity Loss.


\subsection{GaussianFlow} 

We consider the full freedom of each Gaussian motion in a 4D field, including 1) scaling, 2) rotation, and 3) translation at each time step. As the time changes, Gaussians covering the queried pixel at $t=t_1$ will move to other places at $t=t_2$, as shown in Fig.~\ref{fig:example}. To specify new pixel location $\mathbf{x}_{t_2}$ at $t=t_2$, we first project all the 3D Gaussians into 2D image plane as 2D Gaussians and calculate their motion's influence on pixel shifts.

\subsubsection{Flow from Single Gaussian.} To track pixel shifts (flow) contributed by Gaussian motions, we let the relative position of a pixel in a deforming 2D Gaussian stay the same.
This setting makes the probabilities at queried pixel location in Gaussian coordinate system unchanged at two consecutive time steps. According to Eq.~\ref{eq:weights}, the unchanged probability will grant the pixel with the same radiance and opacity contribution from the 2D Gaussian, albeit the 2D Gaussian is deformed. 

The pixel shift (flow) is the image space distance of the same pixel at two time steps.
We first calculate the pixel shift influenced by a single 2D Gaussian that covers the pixel. We can find a pixel $\mathbf{x}$'s location at $t_2$ by normalizing its image location at $t_1$ to canonical Gaussian space and unnormalizing it to image space at $t_2$: 

1) $normalize$. A pixel $\mathbf{x}_{t_1}$ following $i$-th 2D Gaussian distribution can be written as $\mathbf{x}_{t_1}\sim N(\boldsymbol{\mu}_{i,t_1} \mathbf{\Sigma}_{i,t_1})$. And  in $i$-th Gaussian coordinate system with 2D mean $\boldsymbol{\mu}_{i,t_1}\in\mathbb{R}^{2\times1}$ and 2D covariance matrix $\mathbf{\Sigma}_{i,t_1}\in\mathbb{R}^{2\times2}$.  After normalizing the $i$-th Gaussian into the standard normal distribution, we denote the pixel location in canonical Gaussian space as 
\begin{equation}
 \hat{\mathbf{x}}_{t_1}=\mathbf{B}^{-1}_{i,t_1}(\mathbf{x}_{t_1}-\boldsymbol{\mu}_{i,t_1}),
  \label{eq:ours_1}
\end{equation}
which follows $\mathbf{\Sigma}_{i,t_1} = \mathbf{B}_{i,t_1}\mathbf{B}_{i,t_1}^T$, $\hat{\mathbf{x}}_{t_1}\sim N(\mathbf{0}, \mathbf{I})$ and $\mathbf{I}\in\mathbb{R}^{2\times 2}$ is identity matrix. 

2) $unnormalize$. When $t=t_2$, the new location along with the Gaussian motion denotes $\mathbf{x}_{i,t_2}$ on the image plane.
\begin{align}
    \mathbf{x}_{i,t_2} &= \mathbf{B}_{i,t_2}\hat{\mathbf{x}}_{t_1} + \boldsymbol{\mu}_{i,t_2}
     \label{eq:ours_2},
\end{align}
and $\mathbf{\Sigma}_{i,t_2} = \mathbf{B}_{i,t_2}\mathbf{B}_{i,t_2}^T$,  $\mathbf{x}_{t_2}\sim N(\boldsymbol{\mu}_{i,t_2}, \mathbf{\Sigma}_{i,t_2})$. Eq.~\ref{eq:ours_1} and Eq.~\ref{eq:ours_2} preserve Mahalanobis distance between the tracked pixel and the 2D Gaussian leading to consistent probability density across consecutive time steps.
The pixel shift (flow) contribution from each Gaussian therefore can be calculated as:
\begin{align}
    flow^{G}_{i,t_1t_2} = \mathbf{x}_{i,t_2}-\mathbf{x}_{t_1}
\end{align}
\subsubsection{Flow Composition.}
In original 3D Gaussian Splatting, a pixel's color is the weighted sum of the 2D Gaussians' radiance contribution. Similarly, we define the Gaussian flow value at a pixel as the weighted sum of the 2D Gaussians' contributions to its pixel shift, following alpha composition. With Eq.~\ref{eq:ours_1} and Eq.~\ref{eq:ours_2}, the Gaussian flow value at pixel $\mathbf{x}_{t_1}$ from $t=t_{t_1}$ to $t=t_{t_2}$ is 
\begin{align}
   flow^{G}_{t_1t_2} &= \sum^{K}_{i=1} w_iflow^{G}_{i,t_1t_2}\\
   &=\sum^{K}_{i=1}w_i(\mathbf{x}_{i,t_2}-\mathbf{x}_{t_1})\\
   &=\sum^{K}_{i=1}w_i\left[\mathbf{B}_{i,t_2}\mathbf{B}^{-1}_{i,t_1}(\mathbf{x}_{t_1}-\boldsymbol{\mu}_{i,t_1}) + \boldsymbol{\mu}_{i,t_2} - \mathbf{x}_{t_1}) \right],
     \label{eq:ours_3}
\end{align}
where $K$ is the number of Gaussians along each camera ray sorted in depth order and each Gaussian has weight $w_i=\frac{T_i\alpha_i}{\Sigma_i T_i\alpha_i}$ according to Eq.~\ref{eq:rendering_eq}, but normalized to [0,1] along each pixel ray.

In some cases~\cite{ling2023align,keetha2023splatam,yugay2023gaussian,matsuki2023gaussian}, each Gaussian is assumed to be isotropic, and its scaling matrix $\mathbf{S}=\sigma\mathbf{I}$, where $\sigma$ is the scaling factor. And its 3D covariance matrix $\mathbf{RS}\mathbf{S}^T\mathbf{R}^T=\sigma^2\mathbf{I}$. If the scaling factor of each Gaussian doesn't change too much across time, $\mathbf{B}_{i,t_2}\mathbf{B}^{-1}_{i,t_1}\approx \mathbf{I}$. Therefore, to pair with this line of work, the formulation of our Gaussian flow as in Eq.~\ref{eq:ours_3} can be simplified as 
\begin{align}
   flow^{G}_{t_1t_2} &=\sum^{K}_{i=1}w_i(\boldsymbol{\mu}_{i,t_2} - \boldsymbol{\mu}_{i,t_1}).
     \label{eq:ours_4}
\end{align}
In other words, for isotropic Gaussian fields, Gaussian flow between two different time steps can be approximated as the weighted sum of individual translation of 2D Gaussian.

Following either Eq.~\ref{eq:ours_3} or Eq.~\ref{eq:ours_4}, the Gaussian flow can be densely calculated at each pixel. The flow supervision at pixel $\mathbf{x}_{t_1}$ from $t=t_1$ to $t=t_2$ can then be specified as
\begin{align}
   \mathcal{L}_{flow} = ||flow^{o}_{t_1t_2}(\mathbf{x}_{t_1}) - flow^{G}_{t_1t_2}||,
     \label{eq:flow_loss}
\end{align}
where optical flow $flow^{o}_{t_1t_2}$ can be calculated by off-the-shelf methods as pseudo ground-truth.

\begin{figure}[tb]
  \centering
  \includegraphics[height=6.1cm]{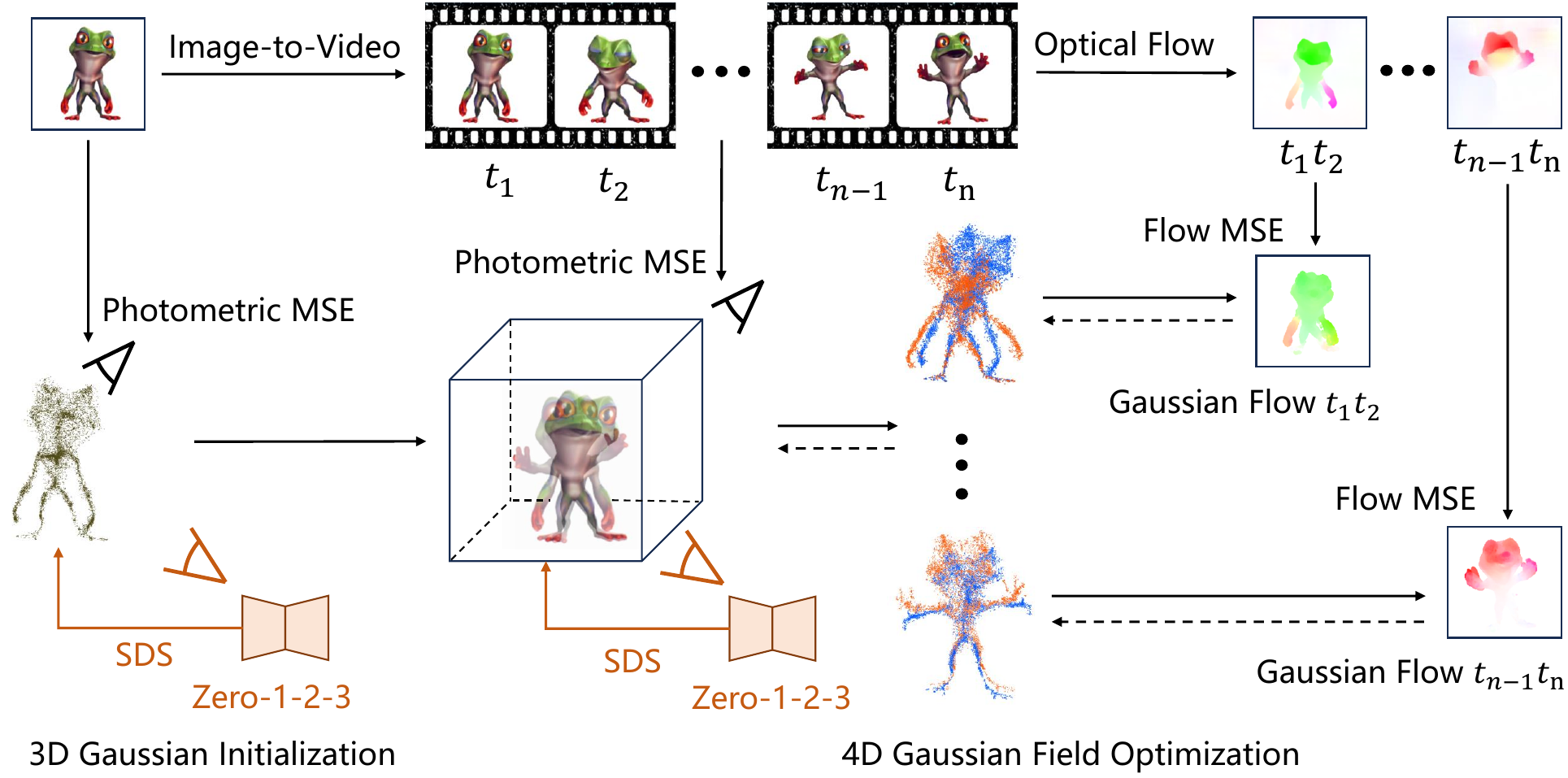}
  \caption{Overview of our 4D content generation pipeline. Our model can take an uncalibrated monocular video or video generated from an image as the input. We optimize a 3D Gaussian field by matching the first frame photometrically on reference view and using a 3D-aware SDS loss \cite{liu2023zero} to supervise the field on novel views. Then, we optimize the dynamics of the 3D Gaussians with the same two losses for each frame. Most importantly, we calculate Gaussian flows on reference view for each consecutive two time step and match it with pre-computed optical flow of the input video. The gradients from the flow matching will propagate back through dynamics splatting and rendering process, resulting in a 4D Gaussian field with natural and smooth motions. 
  }
  \label{fig:overview}
\end{figure}

\subsection{4D Content Generation}
As shown in Fig.~\ref{fig:overview}, 4D content generation with Gaussian representation takes an uncalibrated monocular video either by real capturing or generating from text-to-video or image-to-video models as input and output a 4D Gaussian field.
3D Gaussians are initialized from the first video frame with photometric supervision between rendered image and input image and a 3D-aware diffusion model~\cite{liu2023zero} for multi-view SDS supervision. In our method, 3D Gaussian initialization can be done by One-2-3-45~\cite{liu2024one} or DreamGaussian~\cite{tang2023dreamgaussian}. After initialization, 4D Gaussian field is optimized with per-frame photometric supervision, per-frame SDS supervision, and our flow supervision as in Eq.~\ref{eq:flow_loss}. The loss function for 4D Gaussian field optimization can be written as:
\vspace{-2pt}
\begin{align}
   \mathcal{L} =  \mathcal{L}_{photometric} + \lambda_1\mathcal{L}_{flow} + \lambda_2 \mathcal{L}_{sds}+ \lambda_3 \mathcal{L}_{other},
   \label{eq:overall_generation}
\end{align}
where $\lambda_1$, $\lambda_2$ and $\lambda_3$ are hyperparameters. $\mathcal{L}_{other}$ is optional and method-dependent. Though not used in our method, we leave it for completeness.
\subsection{4D novel view Synthesis}
Unlike 4D content generation that has multi-view object-level prior from 3D-aware diffusion model, 4D novel view synthesis takes only multi-view or monocular input video frames for photometric supervision without any scene-level prior. 3D Gaussians are usually initialized by sfm~\cite{snavely2006photo,schonberger2016structure} from input videos. After initialization, 4D Gaussian field is then optimized with per-frame photometric supervision and our flow supervision. We adopt the 4D Gaussian Fields from \cite{yang2023real}. The loss function for 4D Gaussian field optimization can be written as:

\begin{align}
   \mathcal{L} =  \mathcal{L}_{photometric} + \lambda_1\mathcal{L}_{flow} + \lambda_3\mathcal{L}_{other} ,
     \label{eq:overall_nvs}
\end{align}

\section{Experiments}
\label{sec:blind}
In this section, we first provide implementation details of the proposed method and then valid our method on 4D Gaussian representations with (1) 4D generation and (2) 4D novel view synthesis. We test on the Consistent4D Dataset~\cite{jiang2023consistent4d} and the Plenoptic Video Datasets~\cite{li2022neural} for both quantitative and qualitative evaluation. Our method achieves state-of-the-art results on both tasks.
\vspace{-3mm}
\subsection{Implementation Details}
We take $t_2$ as the next timestep of $t_1$ and calculate optical flow between every two neighbor frames in all experiments. In our CUDA implementation of Gaussian dynamics splatting, though the number of Gaussians $K$ along each pixel ray is usually different, we use $K=20$ to balance speed and effectiveness. A larger $K$ means more number of Gaussians and their gradient will be counted through backpropagation. For video frames with size $H\times W\times 3$, we track the motions of Gaussians between every two neighbor timesteps $t_1$ and $t_2$ by maintaining two $H\times W\times K$ tensors to record the indices of top-$K$ Gaussians sorted in depth order, top-$K$ Gaussians' rendered weights $w_i$ for each pixel and an another tensor with size $H\times W\times K \times 2$ denotes the distances between pixel coordinate and 2D Gaussian means $\mathbf{x}_{t_1}-\boldsymbol{\mu}_{i,t_1}$, respectively. Besides, 2D mean $\boldsymbol{\mu}_{i,t_1}$ and 2D covariance matrices $\mathbf{\Sigma}_{i,t_1}$ and $\mathbf{\Sigma}_{i,t_2}$ of each Gaussian at different two timesteps are accessible via camera projection~\cite{kerbl20233d}.

\begin{table*}[t!] 
\centering
\caption{Quantitative comparisons between ours and others on Consistent4D dataset.
}
\label{tab:con4d}
\scalebox{0.6}
{\begin{tabular}{lcccccccccccccccc}
\toprule
\multirow{2}[2]{*}{Method}  & \multicolumn{2}{c}{\textbf{Pistol}} & \multicolumn{2}{c}{\textbf{Guppie}}& \multicolumn{2}{c}{\textbf{Crocodile}}& \multicolumn{2}{c}{\textbf{Monster}}& \multicolumn{2}{c}{\textbf{Skull}} & \multicolumn{2}{c}{\textbf{Trump}} & \multicolumn{2}{c}{\textbf{Aurorus}} & \multicolumn{2}{c}{\textbf{Mean}} \\ \cmidrule(lr){2-3} \cmidrule(lr){4-5} \cmidrule(lr){6-7} \cmidrule(lr){8-9} \cmidrule(lr){10-11} \cmidrule(lr){12-13}\cmidrule(lr){14-15}\cmidrule(lr){16-17}
 &  LPIPS$\downarrow$  & CLIP$\uparrow$  & LPIPS$\downarrow$  & CLIP$\uparrow$ & LPIPS$\downarrow$  & CLIP$\uparrow$ & LPIPS$\downarrow$  & CLIP$\uparrow$ & LPIPS$\downarrow$  & CLIP$\uparrow$ & LPIPS$\downarrow$  & CLIP$\uparrow$ & LPIPS$\downarrow$  & CLIP$\uparrow$  & LPIPS$\downarrow$ &CLIP$\uparrow$\\
\midrule
D-NeRF~\cite{pumarola2021d}  &  0.52 & 0.66 & 0.32 & 0.76 & 0.54 & 0.61 & 0.52 & 0.79 & 0.53 & 0.72 & 0.55 & 0.60 & 0.56 & 0.66 & 0.51 & 0.68  \\
K-planes~\cite{fridovich2023k}  & 0.40 & 0.74 & 0.29 & 0.75 & 0.19 & 0.75 & 0.47 & 0.73 & 0.41 & 0.72 &   0.51 & 0.66 & 0.37 & 0.67 & 0.38 & 0.72 \\
Consistent4D~\cite{jiang2023consistent4d}  & \textbf{0.10} & 0.90 & 0.12 & 0.90 & 0.12 & 0.82 & 0.18 & 0.90 & \textbf{0.17} & 0.88 & 0.23 & \textbf{0.85} & 0.17 & 0.85 & 0.16 & 0.87 \\
DG4D~\cite{ren2023dreamgaussian4d}  & 0.12 & 0.92 & 0.12  & 0.91  &  0.12 & 0.88 & 0.19 & 0.90 & 0.18 & 0.90 & 0.22 & 0.83 & 0.17 & 0.86 & 0.16 & 0.87 \\
Ours  &  \textbf{0.10} & \textbf{0.94} & \textbf{0.10} & \textbf{0.93} &  \textbf{0.10} & \textbf{0.90} & \textbf{0.17} & \textbf{0.92} & \textbf{0.17} & \textbf{0.92} & \textbf{0.20} & \textbf{0.85} & \textbf{0.15} & \textbf{0.89} & \textbf{0.14} & \textbf{0.91} \\ 

\bottomrule
\end{tabular}}
\vspace{-10pt}
\end{table*}

\begin{figure}[tb]
  \centering
  \includegraphics[height=7.3cm]{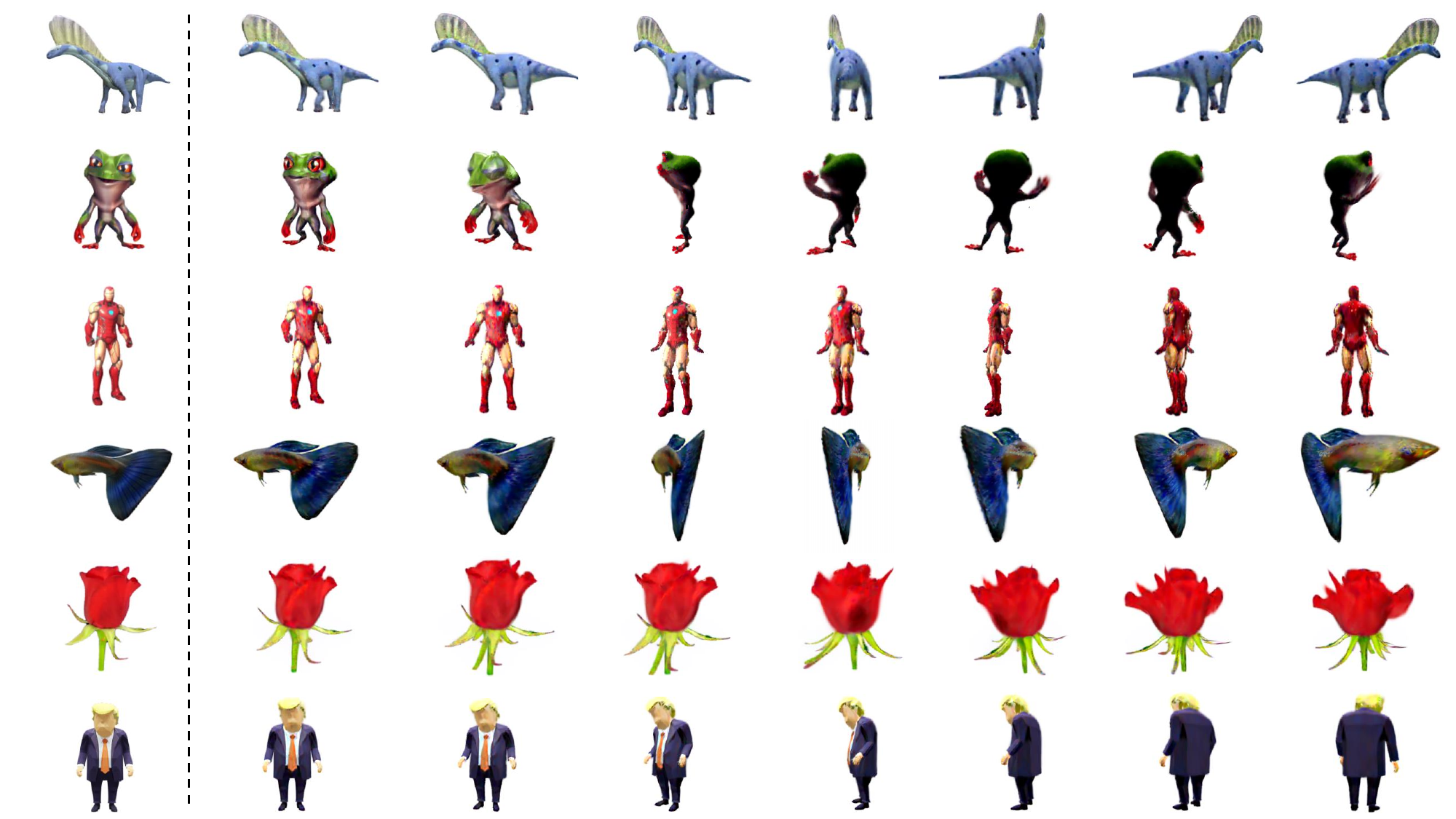}
  \caption{Qualitative results on Consistent4D dataset.
  }
  \label{fig:vis_con4d}
\end{figure}


\begin{figure}[!h]
  \centering
  \includegraphics[height=4.2cm]{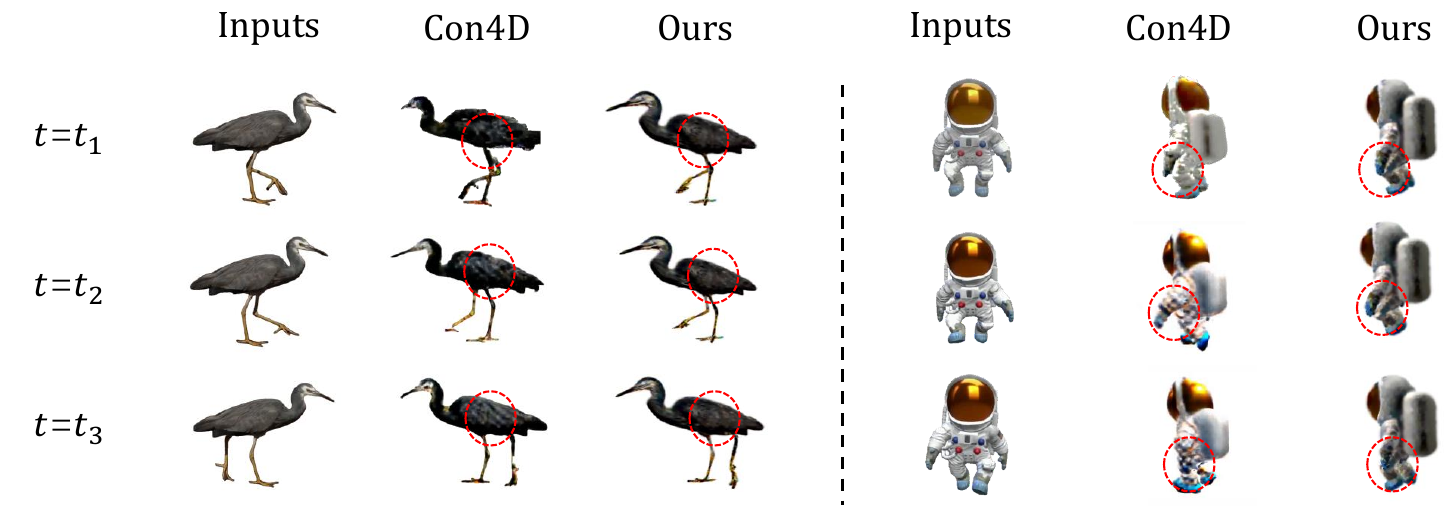}
  \caption{Qualitative comparisons between Consistent4D~\cite{jiang2023consistent4d} (Con4D) and ours. As a dynamic NeRF-based method, Consistent4D shows ``bubble like'' texture and non-consistent geometry on novel views.
  }
  \label{fig:comp_con4d}
\end{figure}

\begin{figure}[!h]
  \centering
  \includegraphics[height=6.6cm]{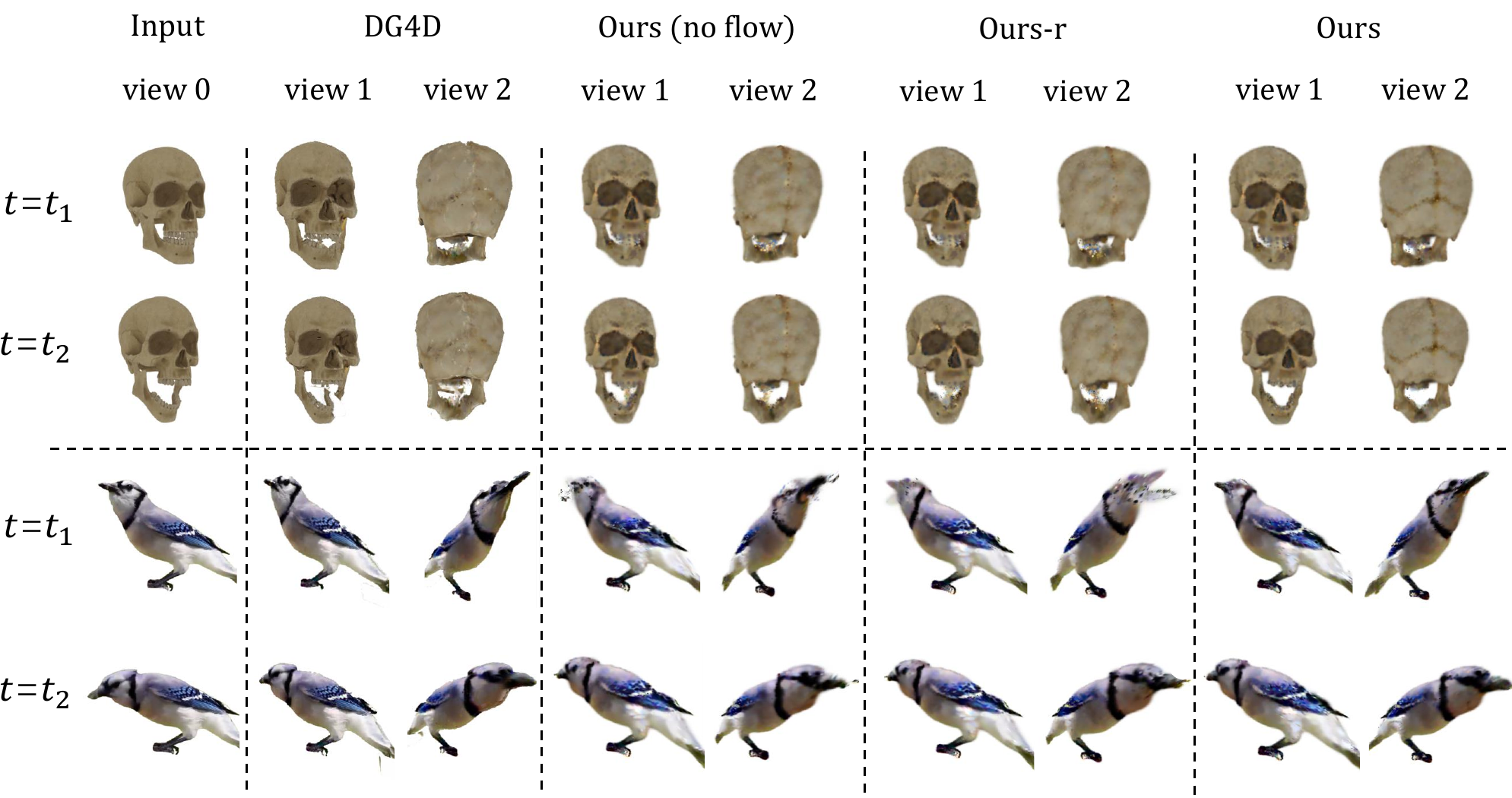}
  \caption{Qualitative comparisons among DreamGaussian4D~\cite{ren2023dreamgaussian4d}, our method without flow loss, our method without flow loss but with Local Rigidity Loss (Ours-r) and ours. 
  }
  \label{fig:failure}
\end{figure}

\vspace{-2mm}
\subsection{Dataset}
\subsubsection{Consistent4D Dataset.} This dataset includes 14 synthetic and 12 in-the-wild monocular videos. All the videos have only one moving object with a white background. 7 of the synthetic videos are provided with multi-view ground-truth for quantitative evaluation. Each input monocular video with a static camera is set at an azimuth angle of 0$^\circ$. Ground-truth images include four distinct views at azimuth angles of -75$^\circ$, 15$^\circ$, 105$^\circ$, and 195$^\circ$, respectively, while keeping elevation, radius, and other camera parameters the same with input camera. 
\vspace{-4mm}
\subsubsection{Plenoptic Video Dataset.} A high-quality real-world dataset consists of 6 scenes with 30FPS and 2028 × 2704 resolution. There are 15 to 20 camera views per scene for training and 1 camera view for testing. Though the dataset has multi-view synchronized cameras, all the viewpoints are mostly limited to the frontal part of scenes.

\subsection{Results and Analysis}
\subsubsection{4D Generation.} We evaluate and compare DreamGaussian4D~\cite{ren2023dreamgaussian4d}, which is a recent 4D Gaussian-based state-of-the-art generative model with open-sourced code, and dynamic NeRF-based methods in Tab.~\ref{tab:con4d} on Consistent4D dataset with ours. Scores on individual videos are calculated and averaged over four novel views mentioned above. Note that flow supervision is effective and helps with 4D generative Gaussian representation. We showcase our superior qualitative results in Fig.~\ref{fig:vis_con4d}. Compared to DreamGaussian4D, our method shows better quality as shown in Fig.~\ref{fig:failure} after the same number of training iterations. 
For the two hard dynamic scenes shown in Fig.~\ref{fig:failure}, our method benefit from flow supervision and generate desirable motions, while DG4D shows prominent artifacts on the novel views. Besides, our method also shows less color drifting compared with dynamic NeRF-based method Consistent4D in Fig.~\ref{fig:comp_con4d}, and our results are more consistent in terms of texture and geometry.

\begin{figure}[!h]
  \centering
  \begin{subfigure}{1.0\textwidth}
      \includegraphics[height=7.2cm]{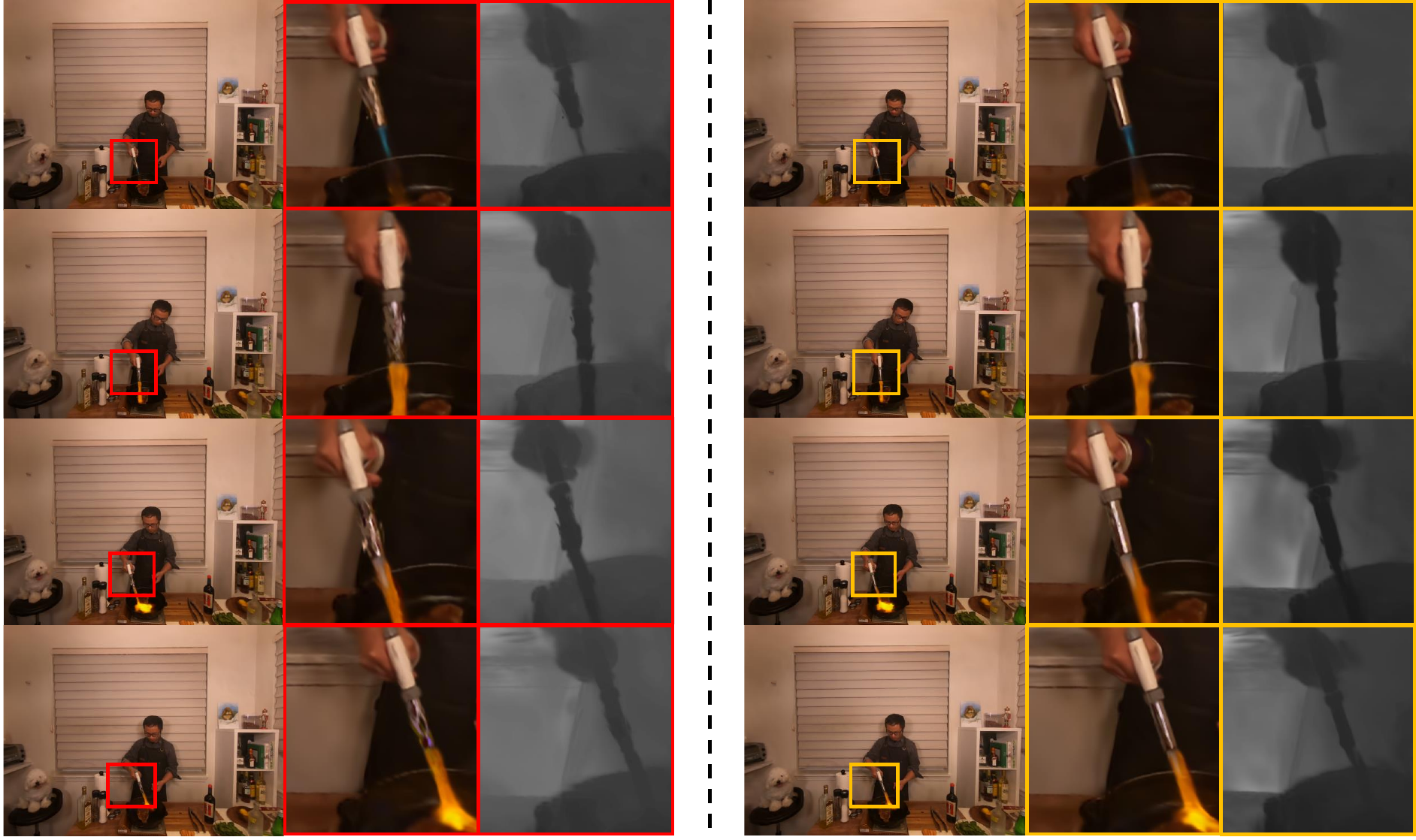}
      \caption{$Flame$ $Steak$ }
      \label{fig:flame_steak}
  \end{subfigure} 
  \par\bigskip
  \begin{subfigure}{1.0\textwidth}
      \includegraphics[height=7.2cm]{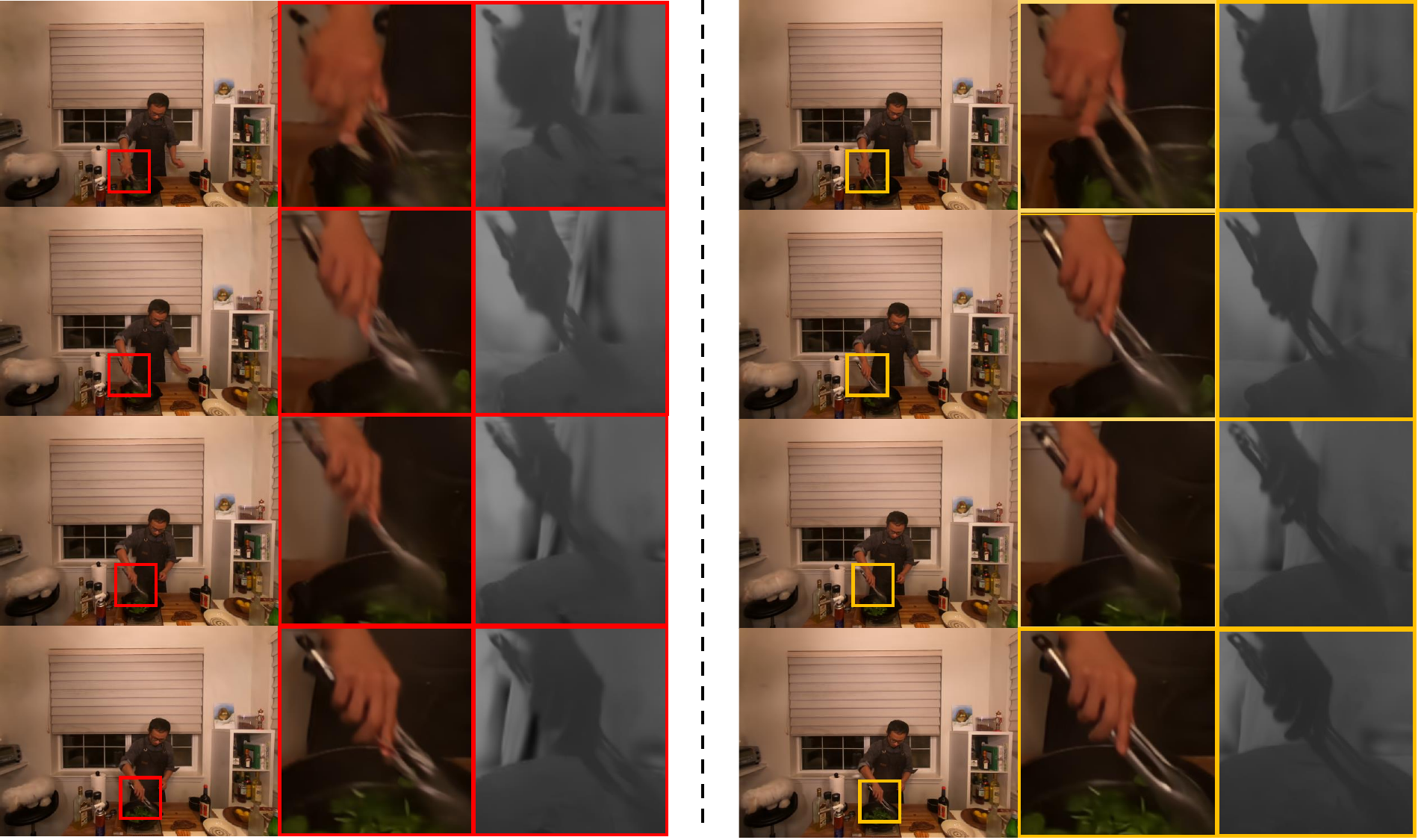}
      \caption{$Cut$ $Spinach$
      }
      \label{fig:Spinach}
  \end{subfigure}%
  \caption{Qualitative comparisons on DyNeRF dataset~\cite{li2022neural}. \textcolor{red}{The 
    left column} shows the novel view rendered images and depth maps of a 4D Gaussian method~\cite{yang2023real}, which suffers from artifacts in the dynamic regions and can hardly handle time-variant specular effect on the moving glossy object. \textcolor{c1}{The right column} shows the results of the same method while optimized with our flow supervision during training. We refer to our supplementary material for more comparisons.
  }
\end{figure}



\vspace{-5mm}
\subsubsection{4D Novel View Synthesis.} We visualize rendered images and depth maps of a very recent state-of-the-art 4D Gaussian method RT-4DGS ~\cite{yang2023real} with (yellow) and without (red) our flow supervision in Fig.~\ref{fig:flame_steak} and Fig.~\ref{fig:Spinach}. According to zoom-in comparisons, our method can consistently model realistic motions and correct structures, even on glossy objects with specular highlights. These regions are known to be challenging~\cite{verbin2022ref,liu2023nero} for most methods, even under adequate multi-view supervision. Our method can reduce ambiguities in photometric supervision by involving motion cues and is shown to be consistently effective across frames. By using an off-the-shelf optical flow algorithm~\cite{shi2023videoflow}, we found that only 1$\%$ to 2$\%$ of image pixels from Plenoptic Video Dataset have optical flow values larger than one pixel. Since our method benefits 4D Gaussian-based methods more on the regions with large motions, we report PSNR numbers on both full scene reconstruction and dynamic regions (optical flow value $>1$) in Tab.~\ref{tab:dynerf_number}. With the proposed flow supervision, our method shows better performance on all scenes and the gains are more prominent on dynamic regions. Consequently, our method also achieves state-of-the art results on 4D novel view synthesis.

\begin{table}[h!]
\centering
\vspace{-5mm}
\caption{Quantitative evaluation between ours and other methods on the DyNeRF dataset \cite{li2022neural}. We report PSNR numbers on both full-scene novel view synthesis and dynamic regions where the ground-truth optical flow value is larger than one pixel. ``Ours'' denotes RT-4DGS with the proposed flow supervision.}
\resizebox{\linewidth}{!}{
\begin{tabular}{l|c|c|c|c|c|c|c}
\toprule
Method & Coffee Martini  & Spinach & Cut Beef & Flame Salmon  & Flame Steak   & Sear Steak & Mean \\ \midrule
HexPlane~\cite{cao2023hexplane}    &  -  & 32.04  &  32.55 & 29.47 & 32.08 & 32.39 & 31.70 \\
K-Planes~\cite{fridovich2023k}  & \textbf{29.99}  & 32.60 & 31.82 & 30.44 & 32.38 & 32.52 & 31.63  \\
MixVoxels~\cite{wang2023mixed}   & 29.36  & 31.61 & 31.30 & 29.92 & 31.21 & 31.43 & 30.80 \\
NeRFPlayer~\cite{song2023nerfplayer}   & 31.53 & 30.56 & 29.35 & \textbf{31.65} & 31.93 & 29.12 & 30.69  \\
HyperReel~\cite{attal2023hyperreel} & 28.37  & 32.30 & 32.92 & 28.26 & 32.20 & 32.57 & 31.10 \\
4DGS~\cite{wu20234d}   & 27.34 & 32.46 & 32.90 & 29.20 & 32.51 & 32.49 & 31.15 \\
RT-4DGS~\cite{yang2023real}      & 28.33 & 32.93 & 33.85 & 29.38 & 34.03 & 33.51 & 32.01 \\
Ours        & 28.42 & \textbf{33.68} & \textbf{34.12} & 29.36 & \textbf{34.22} & \textbf{34.00} & \textbf{32.30}  \\ \midrule
 \multicolumn{8}{c}{Dynamic Region Only} \\ \midrule
RT-4DGS~\cite{yang2023real} & 27.36 & 27.47 & 34.48 & 23.16 & 26.04 & 29.52 & 28.00\\
Ours & \textbf{28.02}& \textbf{28.71} & \textbf{35.16} & \textbf{23.36} & \textbf{27.53} & \textbf{31.15} & \textbf{28.99}\\
\bottomrule
\end{tabular}
}
\label{tab:dynerf_number}
\vspace{-5mm}
\end{table}

\begin{figure}[h]
  \centering
  \includegraphics[height=6.1cm]{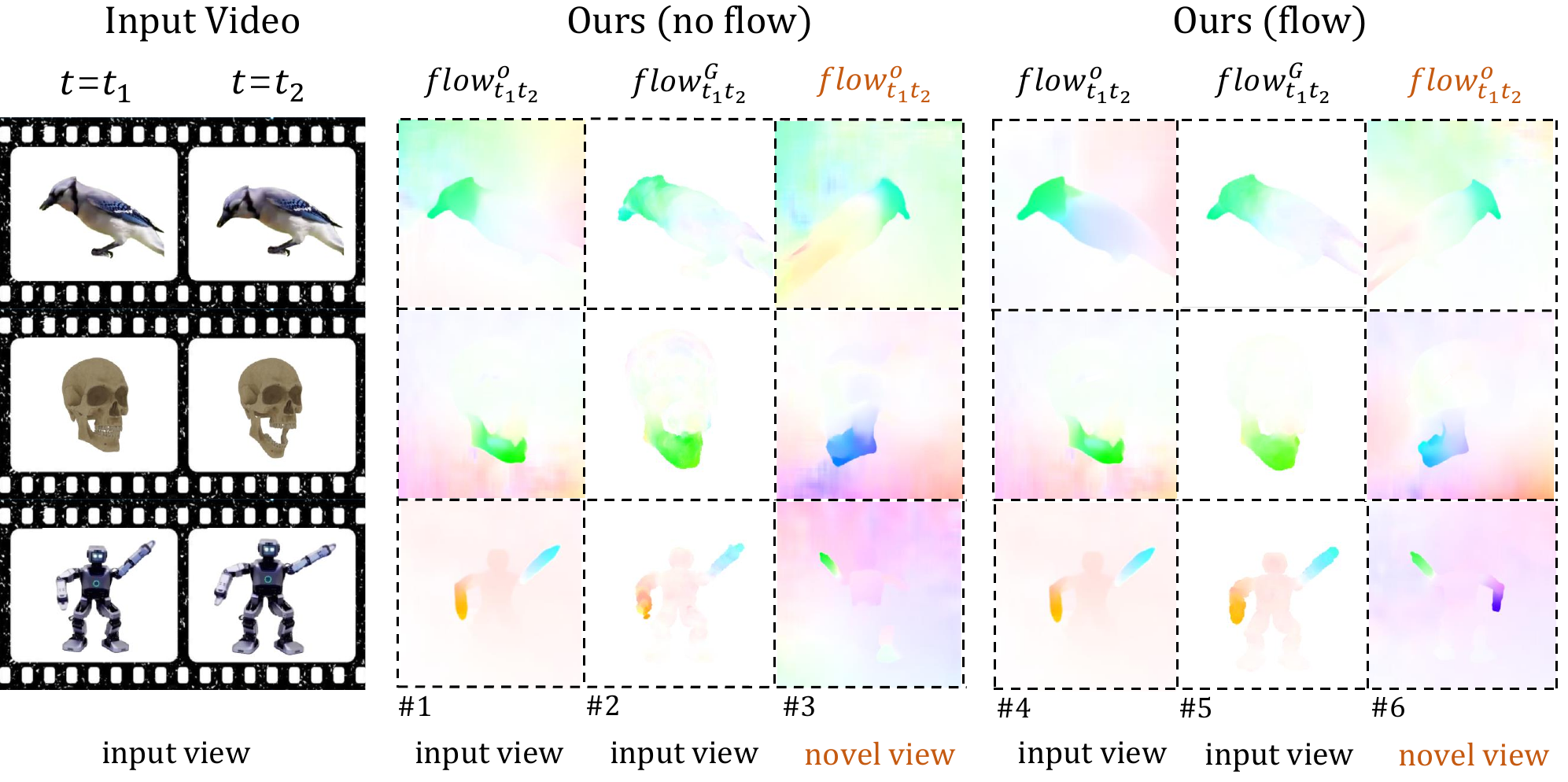}
  \caption{Visualization of optical and Gaussian flows on the input view and a novel view. ``Ours (no flow)'' denotes our model without flow supervision while ``Ours'' is our full model. Note that optical flow values of the background should be ignored because dense optical flow algorithms calculate correspondences among background pixels. We calculate optical flow $flow^o_{t_1t_2}$ on rendered sequences by autoflow~\cite{sun2021autoflow}. From the $\#$1 and the $\#$4 column, we can see that both rendered sequences on input view have high-quality optical flow, indicating correct motions and appearance. Comparing Gaussian flows at the $\#$2 and the $\#$5 column, we can see that the underlining Gaussians will move inconsistently without flow supervision. It is due to the ambiguity of appearance and motions while only being optimized by photometric loss on a single input view. Aligning Gaussian flow to optical flow can drastically improve irregular motions ( $\#$3 column) and create high-quality dynamic motions ($\#$6 column) on novel views. 
  }
  \label{fig:flow_comp}
  \vspace{-3mm}
\end{figure}

\vspace{-3mm}
\section{Ablation Study}
We validate our flow supervision through qualitative comparisons shown in Fig.~\ref{fig:failure}. Compared with Ours (no flow) and Ours, the proposed flow supervision shows its effectiveness on moving parts. For the skull, 3D Gaussians on the teeth region initialized at $t=t_1$ are very close to each other and are hard to split apart completely when $t=t_2$. Because the gradient of incorrectly grouped Gaussians is small due to the small photometric MSE on view 0. Moreover, SDS supervision works on latent domains and cannot provide pixel-wised supervision. And the problem becomes more severe when involving Local Rigidity Loss (comparing Ours-r and Ours) because the motions of 3D Gaussians initialized at $t=t_1$ are constrained by their neighbors and the Gaussians are harder to split apart at $t=t_1$. Similarly, for bird, regions consisting of thin structures such as the bird's beak cannot be perfectly maintained across frames without our flow supervision. While originally utilized in 4D Gaussian fields~\cite{luiten2023dynamic} to maintain the structure consistency during motion, Local Rigidity Loss as a motion constraint can incorrectly group Gaussians and is less effective than our flow supervision. 

We also visualize optical flow $flow^o_{t_1t_2}$ and Gaussian flow $flow^G_{t_1t_2}$ with and without our flow supervision in Fig.~\ref{fig:flow_comp}. In both cases, the optical flow $flow^o_{t_1t_2}$ between rendered images on the input view are very similar to each other (shown in $\#$1 and $\#$ 4 column) and align with ground-truth motion because of direct photometric supervision on input view. However, comparing optical flows on novel view as shown in $\#$3 and $\#$6, without photometric supervision on novel views, inconsistent Gaussian motions are witnessed without our flow supervision. Visualization of Gaussian flow $flow^G_{t_1t_2}$ as in $\#$2 column also reveals the inconsistent Gaussian motions. Incorrect Gaussian motion can still hallucinate correct image frames on input view. However, this motion-appearance ambiguity can lead to unrealistic motions from novel views (the non-smooth flow color on moving parts in $\#$3). While $\#$5 shows consistent Gaussian flow, indicating the consistent Gaussian motions with flow supervision.


\section{Conclusion and Future Work}
We present GaussianFlow, an analytical solution to supervise 3D Gaussian dynamics including scaling, rotation, and translation with 2D optical flow. Extensive qualitative and quantitative comparisons demonstrate that our method is general and beneficial to Gaussian-based representations for both 4D generation and 4D novel view synthesis with motions. 
In this paper, we only consider the short-term flow supervision between every two neighbor frames in our all experiments. Long-term flow supervision across multiple frames is expected to be better and smoother, which we leave as future work. Another promising future direction is to explore view-conditioned flow SDS to supervise Gaussian flow on novel view in the 4D generation task.

\section{Acknowledgments}
We thank Zhengqi Li and Jianchun Chen for thoughtful and valuable discussions.  




%
%
\bibliographystyle{splncs04}
\bibliography{main}

\appendix
\title{Appendix}
\author{}
\institute{}
\maketitle
\input{appendix}


\end{document}

%% file: teaser.tex
\vspace{-1em}
\begin{center}
    \captionsetup{type=figure}
    \includegraphics[width=0.99\linewidth]{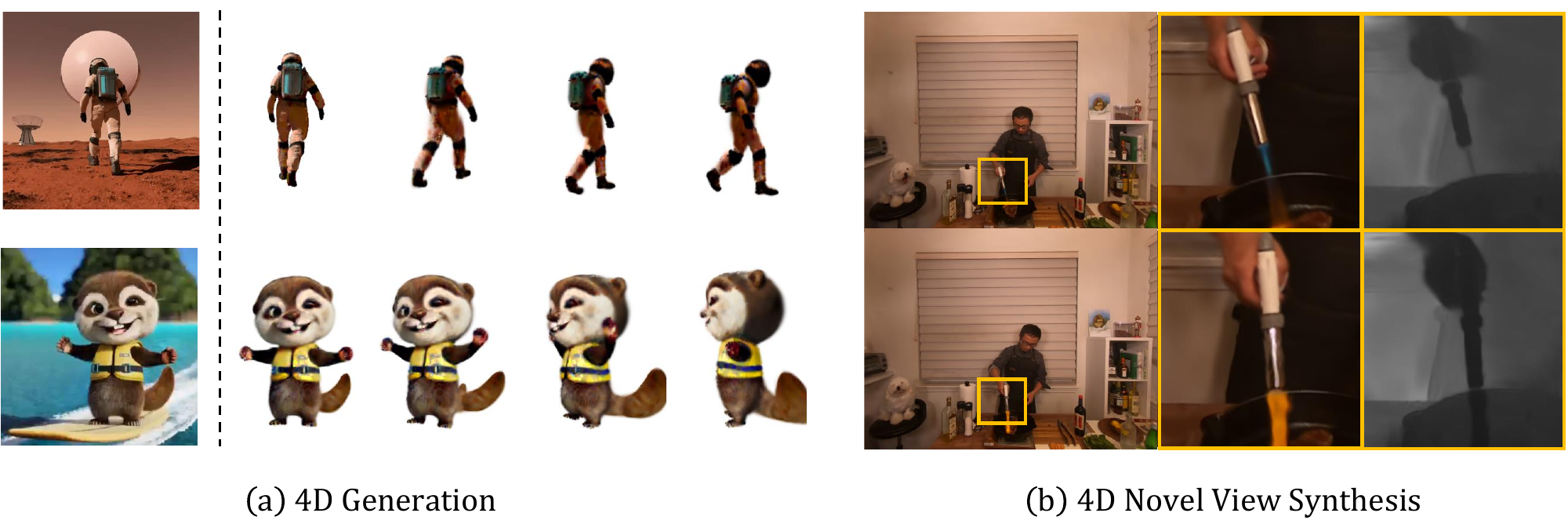}
    \captionof{figure}{We propose Gaussian flow, a dense 2D motion flow created by splatting 3D Gaussian dynamics, which significantly benefits tasks such as 4D generation and 4D novel view synthesis. (a) Based on monocular videos generated by Lumiere \cite{bar2024lumiere} and Sora \cite{videoworldsimulators2024}, our model can generate 4D Gaussian Splatting fields that represent high-quality appearance, geometry and motions. (b) For 4D novel view synthesis, the motions in our generated 4D Gaussian fields are smooth and natural, even in highly dynamic regions where other existing methods suffer from undesirable artifacts.
    }
    \label{fig:teaser}
\end{center}

%% file: appendix.tex
\section{Additional Implementation Details}
A detailed pseudo code for our flow supervision can be found at Algorithm $\color{red}{1}$. We extract the projected Gaussian dynamics and obtain the final Gaussian flow by rendering these dynamics. Variables including the weights and top-$K$ indices of Gaussians per pixel (as mentioned in implementation details of our main paper) are calculated in CUDA by modifying the original CUDA kernel codes of 3D Gaussian Splatting \cite{kerbl20233d}. And Gaussian flow $flow^G$ is calculated by Eq.\textcolor{red}{8} with PyTorch. 

In our 4D generation experiment, we run 500 iterations  static optimization to initialize 3D Gaussian fields with 
a batch size of 16. The Tmax in SDS is linearly decayed  
from 0.98 to 0.02. For dynamic representation, we run
600 iterations with batch size of 4 for both DG4D \cite{ren2023dreamgaussian4d} and ours. The flow loss weight $\lambda_1$ in Eq. \textcolor{red}{11} of our main paper is $1.0$.

In our 4D novel view synthesis experiment, we follow RT-4DGS\cite{yang2023real} except that we add our proposed flow supervision for all cameras. The flow loss weight $\lambda_1$ in Eq. 11 of our main paper is $0.5$.  

\begin{algorithm}[h]
\label{alg1}
\caption{Detailed pseudo code for GaussianFlow}
\KwIn{ \\
$flow^{o}_{t_k,t_{k+1}}$ : Pseudo ground-truth optical flow from off-the-shelf optical flow algorithm; \\ $I^{gt}_{t_k}$: ground-truth images , where $k=0,1,..., T$; \\ $renderer$: A Gaussian renderer; \\  $Gaussians_{t_{k}}$, $Gaussians_{t_{k+1}}$ : $n$ Gaussians with learnable parameters at $t_{k}$ and $t_{k+1}$; \\
$cam_{t_k}$ and $cam_{t_{k+1}}$: Camera parameters at  $t_{k}$ and $t_{k+1}$;}
\textcolor{c2}{\# Loss init}\\
$\mathcal{L}$ = 0\\
\For{{\rm timestep $k\leq T-1$}}{
    // renderer outputs at $t_k$\\
    $renderer_{t_k} = renderer(Gaussians_{t_k}, cam_{t_k})$;\\
    $I^{render}_{t_k} = renderer_{t_k}\left[``image" \right]$; $\quad$ \textcolor{c2}{\#  $H\times W\times 3$} \\ 
    $idx_{t_k} = renderer_{t_k}\left[``index" \right]$; $\quad$ \textcolor{c2}{\# $H\times W\times K $, Gaussian indices that cover each pixels}\\
    $w_{t_k} = renderer_{t_k}\left[``weights" \right]$; $\quad$\textcolor{c2}{\# $H\times W\times K $}\\
    $w_{t_k} = w_{t_k}/sum(w_{t_k}, dim=-1)$; $\quad$\textcolor{c2}{\# $H\times W\times K $, weight normalization}\\ 
    $x\_\mu_{t_k} = renderer_{t_k}\left[``x\_mu" \right]$; \textcolor{c2}{\# $H\times W\times K \times 2, denotes \quad x_{t_k} - \mu_{t_k}  $}\\
    $\mu_{t_k} = renderer_{t_k}\left[``2D\_mean" \right]$; \textcolor{c2}{\# $n\times 2 $}\\ 
    $\Sigma_{t_k} = renderer_{t_k}\left[``2D\_cov" \right]$;   $\quad$ \textcolor{c2}{\# $n\times 2 \times2$} \\
    $B_{t_k} = \Sigma_{t_k}^{\frac{1}{2}}$; \\
    \textcolor{c2}{\# renderer outputs at $t_{k+1}$}\\
    $renderer_{t_{k+1}} = renderer(Gaussians_{t_{k+1}}, cam_{t_{k+1}})$;\\
    $\mu_{t_{k+1}} = renderer_{t_{k+1}}\left[``2D\_mean" \right]$; \textcolor{c2}{\# $n\times 2 $}\\ 
    $\Sigma_{t_{k+1}} = renderer_{t_{k+1}}\left[``2D\_cov" \right]$;$\quad$ 
    \textcolor{c2}{\# $n \times 2 \times2$} \\
    $B_{t_{k+1}} = \Sigma_{t_{k+1}}^{\frac{1}{2}}$; \\
     \textcolor{c2}{\# Eq.8 while ignoring resize operations for simplicity}\\
     $flow^G_{t_k,t_{k+1}} = w_{t_k}*\left(B_{t_{k+1}}[idx_{t_k}]*inv(B_{t_{k}})[idx_{t_k}]*x\_\mu_{t_k} + (\mu_{t_{k+1}}[idx_{t_k}]-\mu_{t_{k}}[idx_{t_k}]-x\_\mu_{t_k}) \right)$\\
     \textcolor{c2}{\# Eq.10} \\
    $\mathcal{L}_{flow} = norm(flow^o_{t_k,t_{k+1}}, sum(flow^G_{t_k,t_{k+1}}, dim=0))$\\
     \textcolor{c2}{\# (1) Loss for 4D novel view synthesis}\\
     $\mathcal{L} = \mathcal{L} + \mathcal{L}_{photometric}(I^{render}_{t_k}, I^{gt}_{t_k}) + \lambda_1\mathcal{L}_{flow} + \lambda_3\mathcal{L}_{other} $ \\
    \textcolor{c2}{\# (2) Loss for 4D generation}\\
     $\mathcal{L} = \mathcal{L} + \mathcal{L}_{photometric}(I^{render}_{t_k}, I^{gt}_{t_k})  + \lambda_1\mathcal{L}_{flow} + \lambda_2\mathcal{L}_{sds}+ \lambda_3\mathcal{L}_{other}$ 
    }
\end{algorithm}




\section{More Results}
\subsection{More Gaussian Flow in 4D Generation.} 
\begin{figure}[ht]
  \centering
  \begin{subfigure}{1.0\textwidth}
      \includegraphics[height=7.1cm]{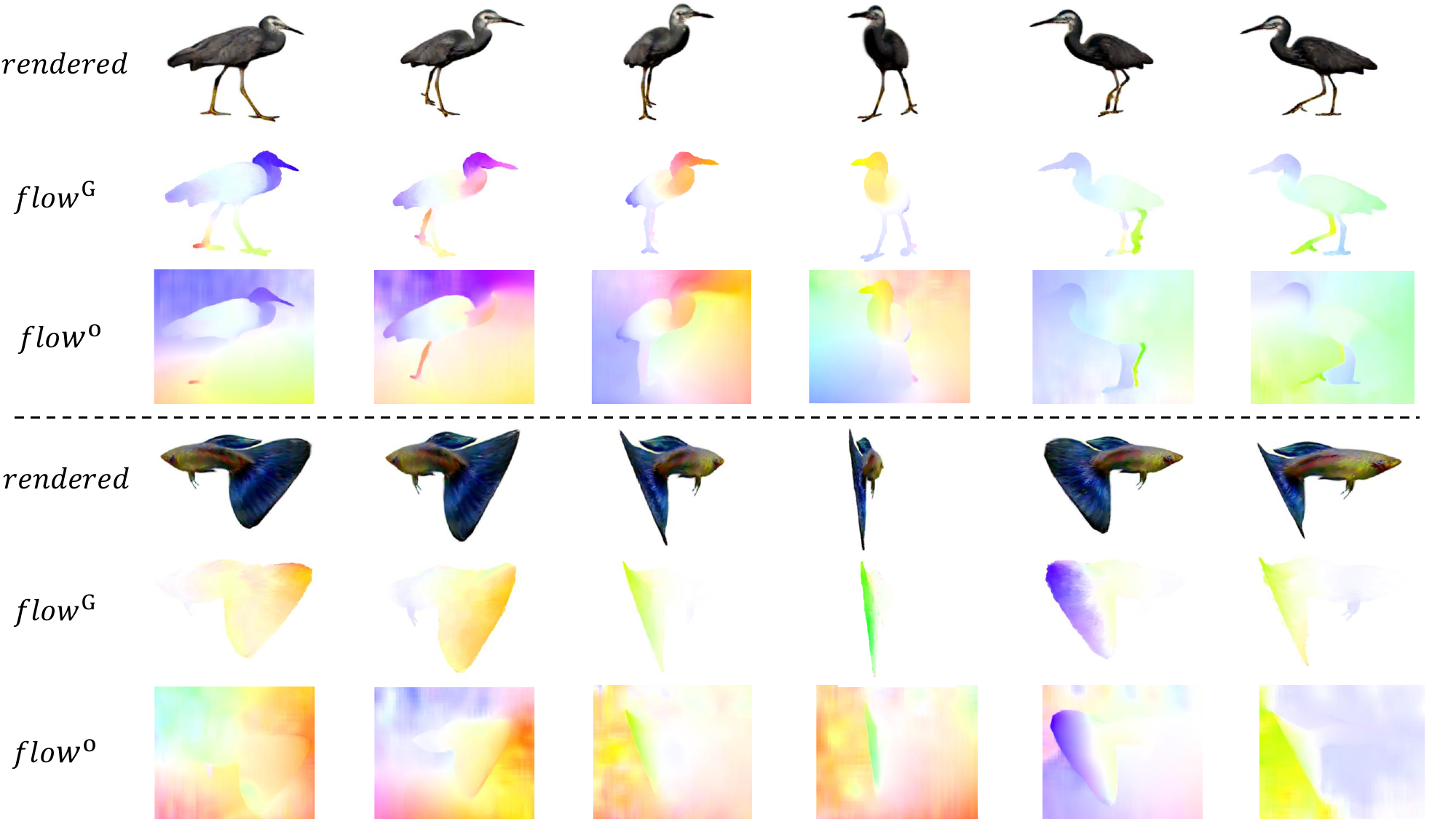}
  \end{subfigure} 
  \par\bigskip
  \begin{subfigure}{0.97\textwidth}
      \includegraphics[height=7.1cm]{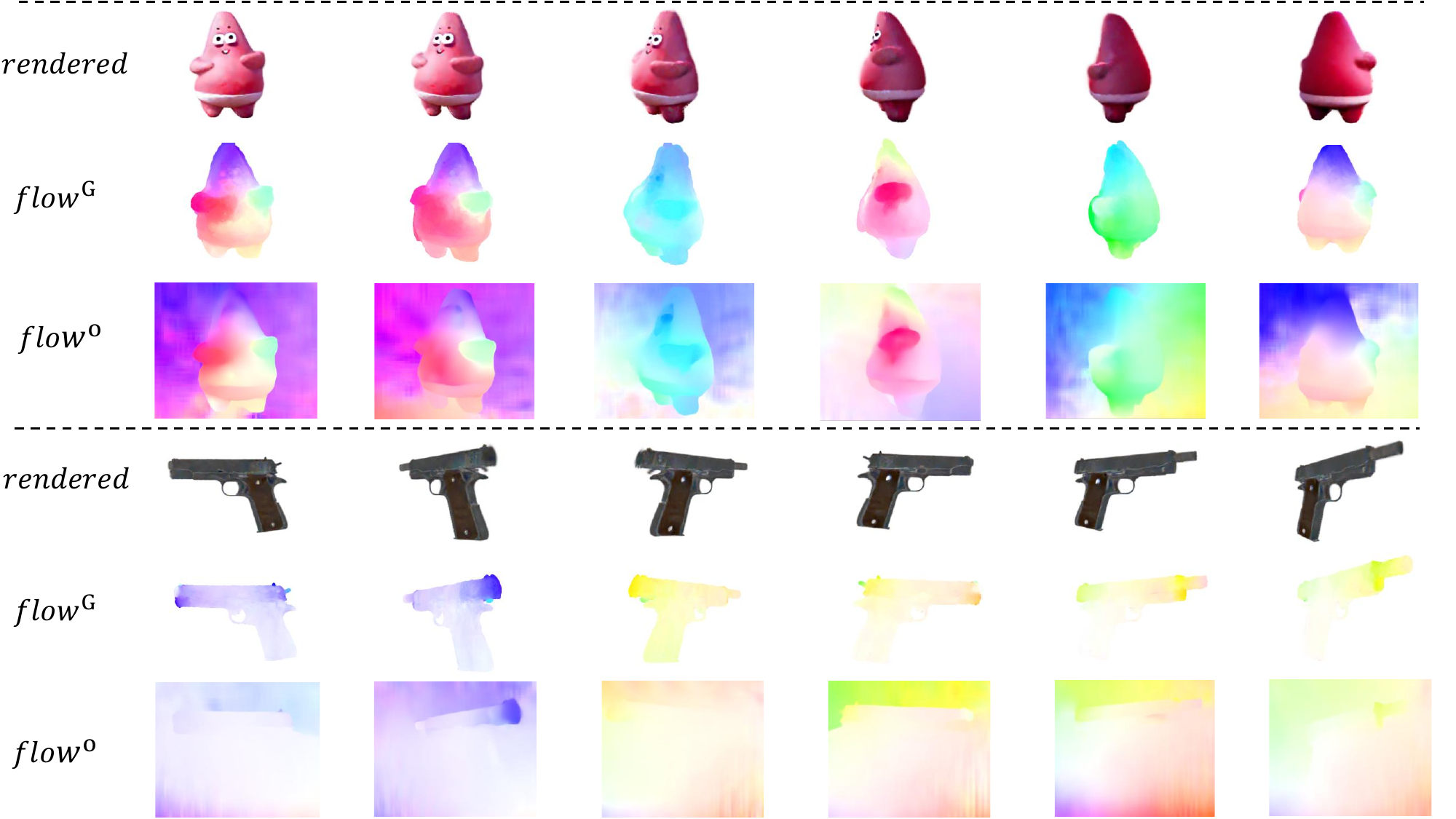}
  \end{subfigure}%
  \caption{Visualization of Gaussian flow $flow^G$ and optical flow $flow^o$ on rendered sequences from different views.
  }
  \label{fig:supp_flow}
\end{figure}

More comparisons between Gaussian flow $flow^G$ and optical flow $flow^o$ on rendered images are shown in Fig.~\ref{fig:supp_flow}. The first row of each example is the rgb frames rendered from a optimized 4D Gaussian field. We rotate our cameras for each time steps so that the object can move as optimized and the camera is moving at the same time to show the scene from different angles. The second row of each example shows the visualized Gaussian flows. These Gaussian flows are calculated by the rendered images of consecutive time steps at each camera view, therefore, containing no camera motion in the flow values. The third row is the estimated optical flows between the rendered images of consecutive time steps at each camera view. We use off-the-shelf AutoFlow \cite{sun2021autoflow} for the estimation. We can see that enhanced by the flow supervision from the single input view, our 4D generation pipeline can model fast motion such as 
the explosive motion of the gun hammer (see the last example in Fig.~\ref{fig:supp_flow}).

\subsection{More Results on the DyNeRF Dataset.} 
\begin{figure}[ht]
  \centering
  \begin{subfigure}{1.0\textwidth}
      \includegraphics[height=7.2cm]{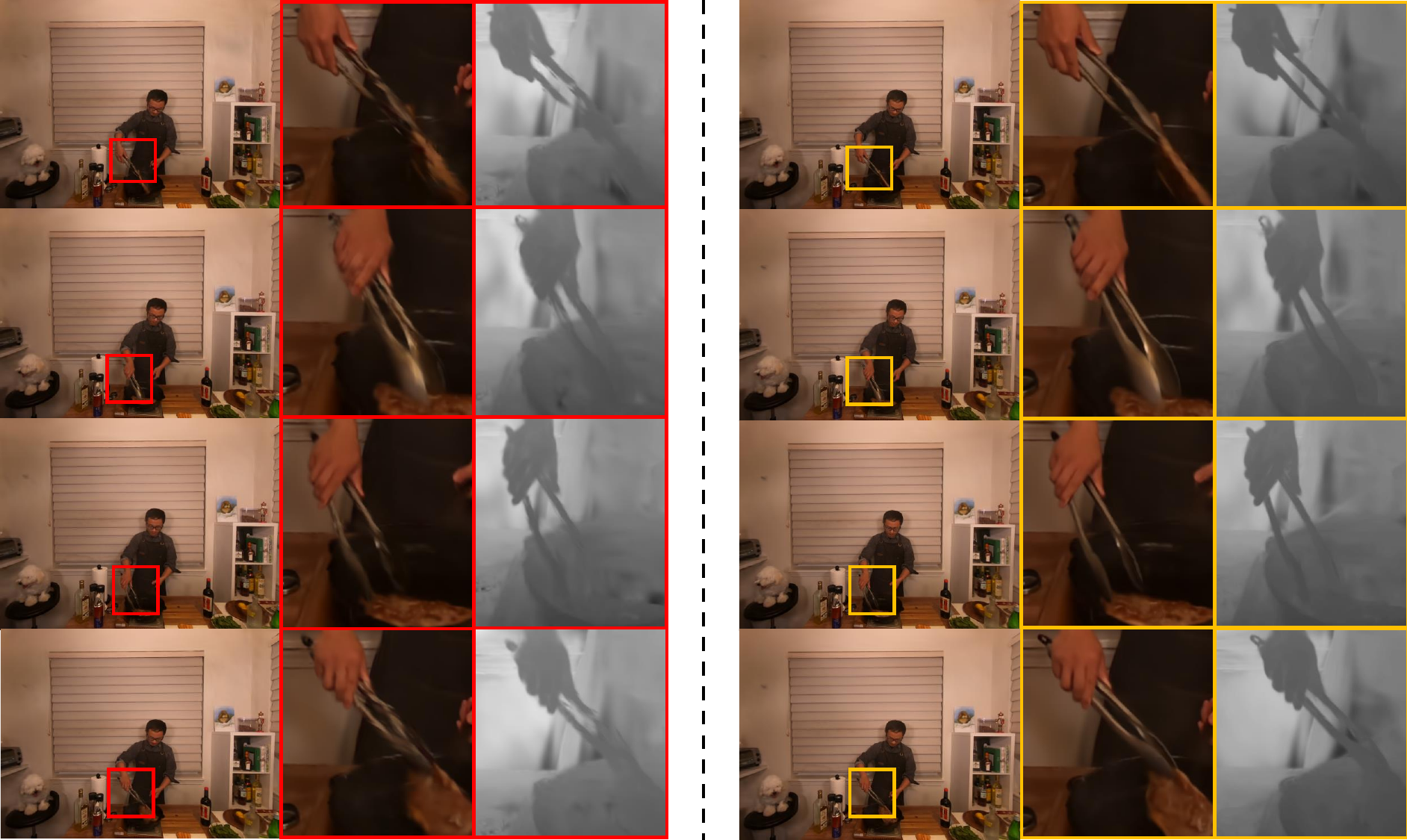}
      \caption{$Sear$ $Steak$ }
  \end{subfigure} 
  \par\bigskip
  \begin{subfigure}{1.0\textwidth}
      \includegraphics[height=7.1cm]{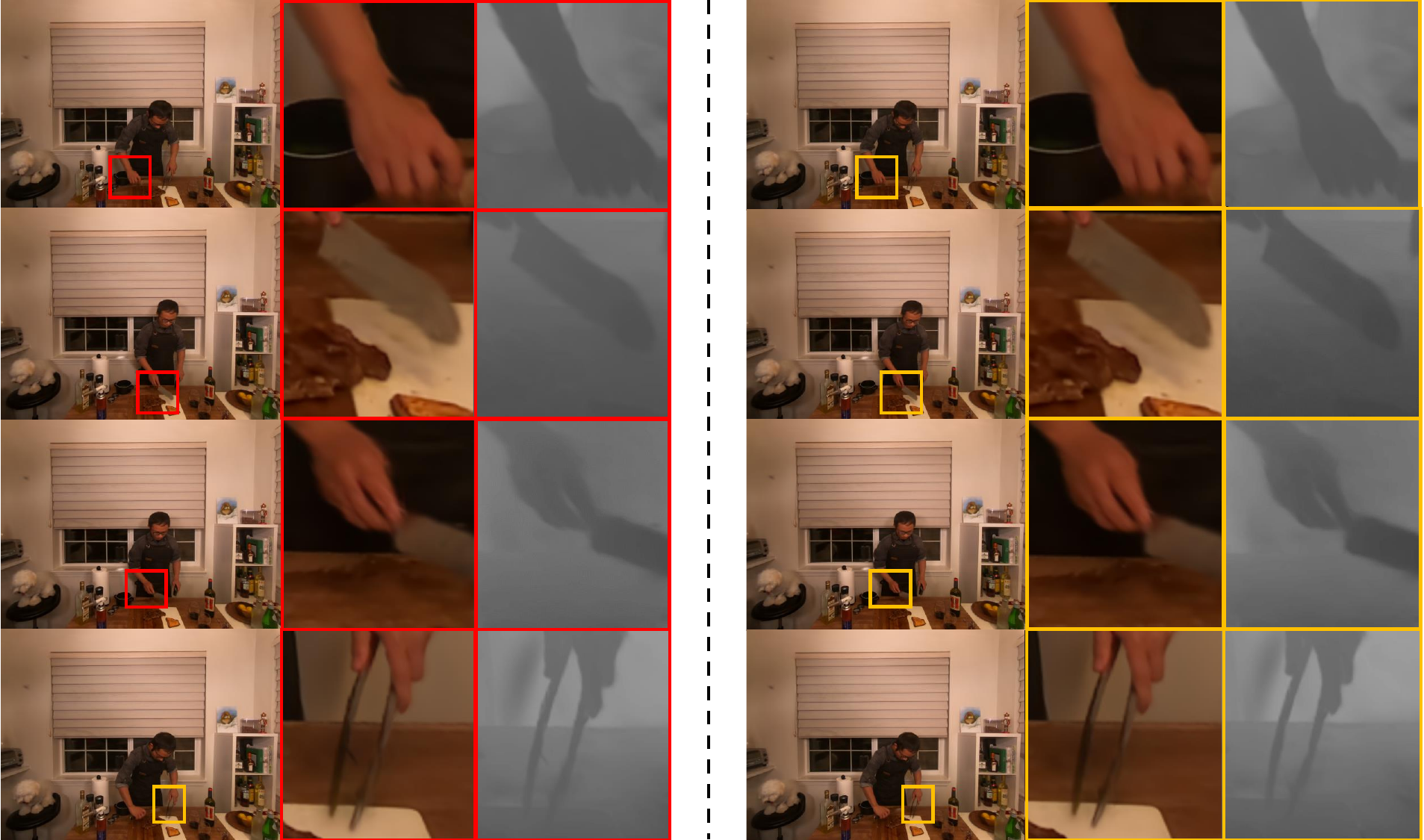}
      \caption{$Cut$ $Beef$}
  \end{subfigure}%
  \caption{Qualitative comparisons on DyNeRF dataset~\cite{li2022neural}. \textcolor{red}{The 
    left column} shows the novel view rendered images and depth maps of a 4D Gaussian method~\cite{yang2023real}. While \textcolor{c1}{The right column} shows the results of the same method while optimized with our flow supervision during training. 
  }
  \label{fig:more_dynerf}
\end{figure}

More qualitative results on DyNeRF dataset~\cite{li2022neural} can be found in Fig.~\ref{fig:more_dynerf} and our video. 

\begin{figure}[h!]
  \centering
      \includegraphics[height=9cm]{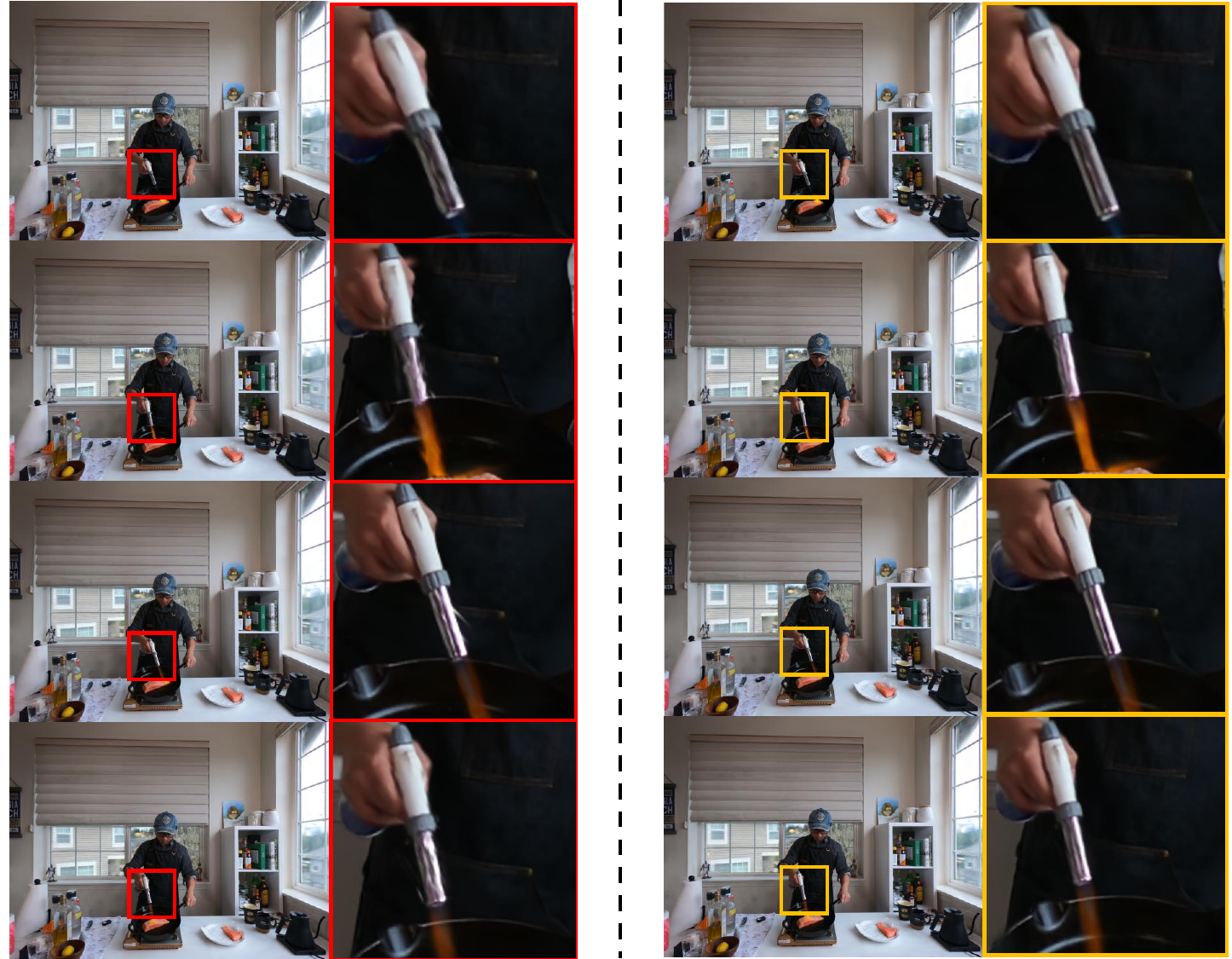}
      \caption{$Flame$ $Salmon$ }
      \label{fig:flame_salmon}
  
  \caption{Qualitative comparisons on DyNeRF dataset~\cite{li2022neural}. Since the details of depth maps on $Flame$ $Salmon$ are hard to be recognized, we only compare the rendered images. \textcolor{red}{The 
    left column} shows the novel view rendered images of a 4D Gaussian method~\cite{yang2023real}. While \textcolor{c1}{The right column} shows the results of the same method while optimized with our flow supervision during training. 
  }
\end{figure}

%
%

%% file: main.bbl
\begin{thebibliography}{10}
\providecommand{\url}[1]{\texttt{#1}}
\providecommand{\urlprefix}{URL }
\providecommand{\doi}[1]{https://doi.org/#1}

\bibitem{attal2023hyperreel}
Attal, B., Huang, J.B., Richardt, C., Zollhoefer, M., Kopf, J., O’Toole, M., Kim, C.: Hyperreel: High-fidelity 6-dof video with ray-conditioned sampling. In: Proceedings of the IEEE/CVF Conference on Computer Vision and Pattern Recognition. pp. 16610--16620 (2023)

\bibitem{bahmani20234d}
Bahmani, S., Skorokhodov, I., Rong, V., Wetzstein, G., Guibas, L., Wonka, P., Tulyakov, S., Park, J.J., Tagliasacchi, A., Lindell, D.B.: 4d-fy: Text-to-4d generation using hybrid score distillation sampling. arXiv preprint arXiv:2311.17984  (2023)

\bibitem{bar2024lumiere}
Bar-Tal, O., Chefer, H., Tov, O., Herrmann, C., Paiss, R., Zada, S., Ephrat, A., Hur, J., Li, Y., Michaeli, T., et~al.: Lumiere: A space-time diffusion model for video generation. arXiv preprint arXiv:2401.12945  (2024)

\bibitem{videoworldsimulators2024}
Brooks, T., Peebles, B., Holmes, C., DePue, W., Guo, Y., Jing, L., Schnurr, D., Taylor, J., Luhman, T., Luhman, E., Ng, C., Wang, R., Ramesh, A.: Video generation models as world simulators  (2024), \url{https://openai.com/research/video-generation-models-as-world-simulators}

\bibitem{cao2023hexplane}
Cao, A., Johnson, J.: Hexplane: A fast representation for dynamic scenes. In: Proceedings of the IEEE/CVF Conference on Computer Vision and Pattern Recognition. pp. 130--141 (2023)

\bibitem{chan2022efficient}
Chan, E.R., Lin, C.Z., Chan, M.A., Nagano, K., Pan, B., De~Mello, S., Gallo, O., Guibas, L.J., Tremblay, J., Khamis, S., et~al.: Efficient geometry-aware 3d generative adversarial networks. In: Proceedings of the IEEE/CVF Conference on Computer Vision and Pattern Recognition. pp. 16123--16133 (2022)

\bibitem{chen2022tensorf}
Chen, A., Xu, Z., Geiger, A., Yu, J., Su, H.: Tensorf: Tensorial radiance fields. In: European Conference on Computer Vision. pp. 333--350. Springer (2022)

\bibitem{deitke2023objaverse}
Deitke, M., Schwenk, D., Salvador, J., Weihs, L., Michel, O., VanderBilt, E., Schmidt, L., Ehsani, K., Kembhavi, A., Farhadi, A.: Objaverse: A universe of annotated 3d objects. In: Proceedings of the IEEE/CVF Conference on Computer Vision and Pattern Recognition. pp. 13142--13153 (2023)

\bibitem{downs2022google}
Downs, L., Francis, A., Koenig, N., Kinman, B., Hickman, R., Reymann, K., McHugh, T.B., Vanhoucke, V.: Google scanned objects: A high-quality dataset of 3d scanned household items. In: 2022 International Conference on Robotics and Automation (ICRA). pp. 2553--2560. IEEE (2022)

\bibitem{fridovich2023k}
Fridovich-Keil, S., Meanti, G., Warburg, F.R., Recht, B., Kanazawa, A.: K-planes: Explicit radiance fields in space, time, and appearance. In: Proceedings of the IEEE/CVF Conference on Computer Vision and Pattern Recognition. pp. 12479--12488 (2023)

\bibitem{gao2021dynamic}
Gao, C., Saraf, A., Kopf, J., Huang, J.B.: Dynamic view synthesis from dynamic monocular video. In: Proceedings of the IEEE/CVF International Conference on Computer Vision. pp. 5712--5721 (2021)

\bibitem{gao2023strivec}
Gao, Q., Xu, Q., Su, H., Neumann, U., Xu, Z.: Strivec: Sparse tri-vector radiance fields. In: Proceedings of the IEEE/CVF International Conference on Computer Vision. pp. 17569--17579 (2023)

\bibitem{hong2023lrm}
Hong, Y., Zhang, K., Gu, J., Bi, S., Zhou, Y., Liu, D., Liu, F., Sunkavalli, K., Bui, T., Tan, H.: Lrm: Large reconstruction model for single image to 3d. arXiv preprint arXiv:2311.04400  (2023)

\bibitem{huang2022real}
Huang, Z., Zhang, T., Heng, W., Shi, B., Zhou, S.: Real-time intermediate flow estimation for video frame interpolation. In: European Conference on Computer Vision. pp. 624--642. Springer (2022)

\bibitem{jiang2023consistent4d}
Jiang, Y., Zhang, L., Gao, J., Hu, W., Yao, Y.: Consistent4d: Consistent 360 $\{$$\backslash$deg$\}$ dynamic object generation from monocular video. arXiv preprint arXiv:2311.02848  (2023)

\bibitem{jun2023shap}
Jun, H., Nichol, A.: Shap-e: Generating conditional 3d implicit functions. arXiv preprint arXiv:2305.02463  (2023)

\bibitem{keetha2023splatam}
Keetha, N., Karhade, J., Jatavallabhula, K.M., Yang, G., Scherer, S., Ramanan, D., Luiten, J.: Splatam: Splat, track \& map 3d gaussians for dense rgb-d slam. arXiv preprint arXiv:2312.02126  (2023)

\bibitem{kerbl20233d}
Kerbl, B., Kopanas, G., Leimk{\"u}hler, T., Drettakis, G.: 3d gaussian splatting for real-time radiance field rendering. ACM Transactions on Graphics  \textbf{42}(4) (2023)

\bibitem{li2022neural}
Li, T., Slavcheva, M., Zollhoefer, M., Green, S., Lassner, C., Kim, C., Schmidt, T., Lovegrove, S., Goesele, M., Newcombe, R., et~al.: Neural 3d video synthesis from multi-view video. In: Proceedings of the IEEE/CVF Conference on Computer Vision and Pattern Recognition. pp. 5521--5531 (2022)

\bibitem{li2021neural}
Li, Z., Niklaus, S., Snavely, N., Wang, O.: Neural scene flow fields for space-time view synthesis of dynamic scenes. In: Proceedings of the IEEE/CVF Conference on Computer Vision and Pattern Recognition. pp. 6498--6508 (2021)

\bibitem{li2023dynibar}
Li, Z., Wang, Q., Cole, F., Tucker, R., Snavely, N.: Dynibar: Neural dynamic image-based rendering. In: Proceedings of the IEEE/CVF Conference on Computer Vision and Pattern Recognition. pp. 4273--4284 (2023)

\bibitem{lin2023magic3d}
Lin, C.H., Gao, J., Tang, L., Takikawa, T., Zeng, X., Huang, X., Kreis, K., Fidler, S., Liu, M.Y., Lin, T.Y.: Magic3d: High-resolution text-to-3d content creation. In: Proceedings of the IEEE/CVF Conference on Computer Vision and Pattern Recognition. pp. 300--309 (2023)

\bibitem{ling2023align}
Ling, H., Kim, S.W., Torralba, A., Fidler, S., Kreis, K.: Align your gaussians: Text-to-4d with dynamic 3d gaussians and composed diffusion models. arXiv preprint arXiv:2312.13763  (2023)

\bibitem{liu2023one}
Liu, M., Shi, R., Chen, L., Zhang, Z., Xu, C., Wei, X., Chen, H., Zeng, C., Gu, J., Su, H.: One-2-3-45++: Fast single image to 3d objects with consistent multi-view generation and 3d diffusion. arXiv preprint arXiv:2311.07885  (2023)

\bibitem{liu2024one}
Liu, M., Xu, C., Jin, H., Chen, L., Varma~T, M., Xu, Z., Su, H.: One-2-3-45: Any single image to 3d mesh in 45 seconds without per-shape optimization. Advances in Neural Information Processing Systems  \textbf{36} (2024)

\bibitem{liu2023zero}
Liu, R., Wu, R., Van~Hoorick, B., Tokmakov, P., Zakharov, S., Vondrick, C.: Zero-1-to-3: Zero-shot one image to 3d object. In: Proceedings of the IEEE/CVF International Conference on Computer Vision. pp. 9298--9309 (2023)

\bibitem{liu2023syncdreamer}
Liu, Y., Lin, C., Zeng, Z., Long, X., Liu, L., Komura, T., Wang, W.: Syncdreamer: Generating multiview-consistent images from a single-view image. arXiv preprint arXiv:2309.03453  (2023)

\bibitem{liu2023nero}
Liu, Y., Wang, P., Lin, C., Long, X., Wang, J., Liu, L., Komura, T., Wang, W.: Nero: Neural geometry and brdf reconstruction of reflective objects from multiview images. arXiv preprint arXiv:2305.17398  (2023)

\bibitem{long2023wonder3d}
Long, X., Guo, Y.C., Lin, C., Liu, Y., Dou, Z., Liu, L., Ma, Y., Zhang, S.H., Habermann, M., Theobalt, C., et~al.: Wonder3d: Single image to 3d using cross-domain diffusion. arXiv preprint arXiv:2310.15008  (2023)

\bibitem{luiten2023dynamic}
Luiten, J., Kopanas, G., Leibe, B., Ramanan, D.: Dynamic 3d gaussians: Tracking by persistent dynamic view synthesis. arXiv preprint arXiv:2308.09713  (2023)

\bibitem{matsuki2023gaussian}
Matsuki, H., Murai, R., Kelly, P.H., Davison, A.J.: Gaussian splatting slam. arXiv preprint arXiv:2312.06741  (2023)

\bibitem{mildenhall2021nerf}
Mildenhall, B., Srinivasan, P.P., Tancik, M., Barron, J.T., Ramamoorthi, R., Ng, R.: Nerf: Representing scenes as neural radiance fields for view synthesis. Communications of the ACM  \textbf{65}(1),  99--106 (2021)

\bibitem{newcombe2015dynamicfusion}
Newcombe, R.A., Fox, D., Seitz, S.M.: Dynamicfusion: Reconstruction and tracking of non-rigid scenes in real-time. In: Proceedings of the IEEE conference on computer vision and pattern recognition. pp. 343--352 (2015)

\bibitem{newcombe2011kinectfusion}
Newcombe, R.A., Izadi, S., Hilliges, O., Molyneaux, D., Kim, D., Davison, A.J., Kohi, P., Shotton, J., Hodges, S., Fitzgibbon, A.: Kinectfusion: Real-time dense surface mapping and tracking. In: 2011 10th IEEE international symposium on mixed and augmented reality. pp. 127--136. Ieee (2011)

\bibitem{nichol2022point}
Nichol, A., Jun, H., Dhariwal, P., Mishkin, P., Chen, M.: Point-e: A system for generating 3d point clouds from complex prompts. arXiv preprint arXiv:2212.08751  (2022)

\bibitem{park2021nerfies}
Park, K., Sinha, U., Barron, J.T., Bouaziz, S., Goldman, D.B., Seitz, S.M., Martin-Brualla, R.: Nerfies: Deformable neural radiance fields. In: Proceedings of the IEEE/CVF International Conference on Computer Vision. pp. 5865--5874 (2021)

\bibitem{park2021hypernerf}
Park, K., Sinha, U., Hedman, P., Barron, J.T., Bouaziz, S., Goldman, D.B., Martin-Brualla, R., Seitz, S.M.: Hypernerf: A higher-dimensional representation for topologically varying neural radiance fields. arXiv preprint arXiv:2106.13228  (2021)

\bibitem{poole2022dreamfusion}
Poole, B., Jain, A., Barron, J.T., Mildenhall, B.: Dreamfusion: Text-to-3d using 2d diffusion. arXiv preprint arXiv:2209.14988  (2022)

\bibitem{pumarola2021d}
Pumarola, A., Corona, E., Pons-Moll, G., Moreno-Noguer, F.: D-nerf: Neural radiance fields for dynamic scenes. In: Proceedings of the IEEE/CVF Conference on Computer Vision and Pattern Recognition. pp. 10318--10327 (2021)

\bibitem{radford2021learning}
Radford, A., Kim, J.W., Hallacy, C., Ramesh, A., Goh, G., Agarwal, S., Sastry, G., Askell, A., Mishkin, P., Clark, J., et~al.: Learning transferable visual models from natural language supervision. In: International conference on machine learning. pp. 8748--8763. PMLR (2021)

\bibitem{raj2023dreambooth3d}
Raj, A., Kaza, S., Poole, B., Niemeyer, M., Ruiz, N., Mildenhall, B., Zada, S., Aberman, K., Rubinstein, M., Barron, J., et~al.: Dreambooth3d: Subject-driven text-to-3d generation. arXiv preprint arXiv:2303.13508  (2023)

\bibitem{ren2023dreamgaussian4d}
Ren, J., Pan, L., Tang, J., Zhang, C., Cao, A., Zeng, G., Liu, Z.: Dreamgaussian4d: Generative 4d gaussian splatting. arXiv preprint arXiv:2312.17142  (2023)

\bibitem{rombach2022high}
Rombach, R., Blattmann, A., Lorenz, D., Esser, P., Ommer, B.: High-resolution image synthesis with latent diffusion models. In: Proceedings of the IEEE/CVF conference on computer vision and pattern recognition. pp. 10684--10695 (2022)

\bibitem{schonberger2016structure}
Schonberger, J.L., Frahm, J.M.: Structure-from-motion revisited. In: Proceedings of the IEEE conference on computer vision and pattern recognition. pp. 4104--4113 (2016)

\bibitem{shao2023tensor4d}
Shao, R., Zheng, Z., Tu, H., Liu, B., Zhang, H., Liu, Y.: Tensor4d: Efficient neural 4d decomposition for high-fidelity dynamic reconstruction and rendering. In: Proceedings of the IEEE/CVF Conference on Computer Vision and Pattern Recognition. pp. 16632--16642 (2023)

\bibitem{shi2023videoflow}
Shi, X., Huang, Z., Bian, W., Li, D., Zhang, M., Cheung, K.C., See, S., Qin, H., Dai, J., Li, H.: Videoflow: Exploiting temporal cues for multi-frame optical flow estimation. arXiv preprint arXiv:2303.08340  (2023)

\bibitem{shi2023mvdream}
Shi, Y., Wang, P., Ye, J., Long, M., Li, K., Yang, X.: Mvdream: Multi-view diffusion for 3d generation. arXiv preprint arXiv:2308.16512  (2023)

\bibitem{singer2023text}
Singer, U., Sheynin, S., Polyak, A., Ashual, O., Makarov, I., Kokkinos, F., Goyal, N., Vedaldi, A., Parikh, D., Johnson, J., et~al.: Text-to-4d dynamic scene generation. arXiv preprint arXiv:2301.11280  (2023)

\bibitem{snavely2006photo}
Snavely, N., Seitz, S.M., Szeliski, R.: Photo tourism: exploring photo collections in 3d. In: ACM siggraph 2006 papers, pp. 835--846 (2006)

\bibitem{song2023nerfplayer}
Song, L., Chen, A., Li, Z., Chen, Z., Chen, L., Yuan, J., Xu, Y., Geiger, A.: Nerfplayer: A streamable dynamic scene representation with decomposed neural radiance fields. IEEE Transactions on Visualization and Computer Graphics  \textbf{29}(5),  2732--2742 (2023)

\bibitem{sun2021autoflow}
Sun, D., Vlasic, D., Herrmann, C., Jampani, V., Krainin, M., Chang, H., Zabih, R., Freeman, W.T., Liu, C.: Autoflow: Learning a better training set for optical flow. In: Proceedings of the IEEE/CVF Conference on Computer Vision and Pattern Recognition. pp. 10093--10102 (2021)

\bibitem{sun2023dreamcraft3d}
Sun, J., Zhang, B., Shao, R., Wang, L., Liu, W., Xie, Z., Liu, Y.: Dreamcraft3d: Hierarchical 3d generation with bootstrapped diffusion prior. arXiv preprint arXiv:2310.16818  (2023)

\bibitem{tang2023dreamgaussian}
Tang, J., Ren, J., Zhou, H., Liu, Z., Zeng, G.: Dreamgaussian: Generative gaussian splatting for efficient 3d content creation. arXiv preprint arXiv:2309.16653  (2023)

\bibitem{tretschk2021non}
Tretschk, E., Tewari, A., Golyanik, V., Zollh{\"o}fer, M., Lassner, C., Theobalt, C.: Non-rigid neural radiance fields: Reconstruction and novel view synthesis of a dynamic scene from monocular video. In: Proceedings of the IEEE/CVF International Conference on Computer Vision. pp. 12959--12970 (2021)

\bibitem{verbin2022ref}
Verbin, D., Hedman, P., Mildenhall, B., Zickler, T., Barron, J.T., Srinivasan, P.P.: Ref-nerf: Structured view-dependent appearance for neural radiance fields. in 2022 ieee. In: CVF Conference on Computer Vision and Pattern Recognition (CVPR). pp. 5481--5490 (2022)

\bibitem{wang2021neural}
Wang, C., Eckart, B., Lucey, S., Gallo, O.: Neural trajectory fields for dynamic novel view synthesis. arXiv preprint arXiv:2105.05994  (2021)

\bibitem{wang2023flow}
Wang, C., MacDonald, L.E., Jeni, L.A., Lucey, S.: Flow supervision for deformable nerf. In: Proceedings of the IEEE/CVF Conference on Computer Vision and Pattern Recognition. pp. 21128--21137 (2023)

\bibitem{wang2023mixed}
Wang, F., Tan, S., Li, X., Tian, Z., Song, Y., Liu, H.: Mixed neural voxels for fast multi-view video synthesis. In: Proceedings of the IEEE/CVF International Conference on Computer Vision. pp. 19706--19716 (2023)

\bibitem{wang2023pf}
Wang, P., Tan, H., Bi, S., Xu, Y., Luan, F., Sunkavalli, K., Wang, W., Xu, Z., Zhang, K.: Pf-lrm: Pose-free large reconstruction model for joint pose and shape prediction. arXiv preprint arXiv:2311.12024  (2023)

\bibitem{wang2024prolificdreamer}
Wang, Z., Lu, C., Wang, Y., Bao, F., Li, C., Su, H., Zhu, J.: Prolificdreamer: High-fidelity and diverse text-to-3d generation with variational score distillation. Advances in Neural Information Processing Systems  \textbf{36} (2024)

\bibitem{wu20234d}
Wu, G., Yi, T., Fang, J., Xie, L., Zhang, X., Wei, W., Liu, W., Tian, Q., Wang, X.: 4d gaussian splatting for real-time dynamic scene rendering. arXiv preprint arXiv:2310.08528  (2023)

\bibitem{xu2023dmv3d}
Xu, Y., Tan, H., Luan, F., Bi, S., Wang, P., Li, J., Shi, Z., Sunkavalli, K., Wetzstein, G., Xu, Z., et~al.: Dmv3d: Denoising multi-view diffusion using 3d large reconstruction model. arXiv preprint arXiv:2311.09217  (2023)

\bibitem{yang2021lasr}
Yang, G., Sun, D., Jampani, V., Vlasic, D., Cole, F., Chang, H., Ramanan, D., Freeman, W.T., Liu, C.: Lasr: Learning articulated shape reconstruction from a monocular video. In: Proceedings of the IEEE/CVF Conference on Computer Vision and Pattern Recognition. pp. 15980--15989 (2021)

\bibitem{yang2021viser}
Yang, G., Sun, D., Jampani, V., Vlasic, D., Cole, F., Liu, C., Ramanan, D.: Viser: Video-specific surface embeddings for articulated 3d shape reconstruction. Advances in Neural Information Processing Systems  \textbf{34},  19326--19338 (2021)

\bibitem{yang2023reconstructing}
Yang, G., Wang, C., Reddy, N.D., Ramanan, D.: Reconstructing animatable categories from videos. In: Proceedings of the IEEE/CVF Conference on Computer Vision and Pattern Recognition. pp. 16995--17005 (2023)

\bibitem{yang2023ppr}
Yang, G., Yang, S., Zhang, J.Z., Manchester, Z., Ramanan, D.: Ppr: Physically plausible reconstruction from monocular videos. In: Proceedings of the IEEE/CVF International Conference on Computer Vision. pp. 3914--3924 (2023)

\bibitem{yang2023real}
Yang, Z., Yang, H., Pan, Z., Zhu, X., Zhang, L.: Real-time photorealistic dynamic scene representation and rendering with 4d gaussian splatting. arXiv preprint arXiv:2310.10642  (2023)

\bibitem{yu2023mvimgnet}
Yu, X., Xu, M., Zhang, Y., Liu, H., Ye, C., Wu, Y., Yan, Z., Zhu, C., Xiong, Z., Liang, T., et~al.: Mvimgnet: A large-scale dataset of multi-view images. In: Proceedings of the IEEE/CVF Conference on Computer Vision and Pattern Recognition. pp. 9150--9161 (2023)

\bibitem{yugay2023gaussian}
Yugay, V., Li, Y., Gevers, T., Oswald, M.R.: Gaussian-slam: Photo-realistic dense slam with gaussian splatting. arXiv preprint arXiv:2312.10070  (2023)

\bibitem{zhao2023animate124}
Zhao, Y., Yan, Z., Xie, E., Hong, L., Li, Z., Lee, G.H.: Animate124: Animating one image to 4d dynamic scene. arXiv preprint arXiv:2311.14603  (2023)

\bibitem{zollhofer2014real}
Zollh{\"o}fer, M., Nie{\ss}ner, M., Izadi, S., Rehmann, C., Zach, C., Fisher, M., Wu, C., Fitzgibbon, A., Loop, C., Theobalt, C., et~al.: Real-time non-rigid reconstruction using an rgb-d camera. ACM Transactions on Graphics (ToG)  \textbf{33}(4),  1--12 (2014)

\end{thebibliography}
